%% file: _main.tex
\input{_constants}
\arxiv %

\pdfoutput=1
\documentclass[10pt,twocolumn,letterpaper]{article}
\input{cvpr_header}

\unless\ifarxiv \myexternaldocument{_supplementary} \fi

\begin{document}
\title{\paperTitle}
\author{\authorBlock}
\maketitle

\input{00_abstract}
\input{01_intro}

\input{02_related}

\input{03_method}

\input{04_experiment}

\input{05_conclusion}

{\small
\bibliographystyle{ieeenat_fullname}
\bibliography{11_references}
}

\input{12_appendix}

\newpage

\end{document}

%% file: _constants.tex
\def\paperID{11360}
\def\confName{ICCV}
\def\confYear{2025}

\def\paperTitle{An Efficient Post-hoc Framework for Reducing Task Discrepancy of\\Text Encoders for Composed Image Retrieval}

\def\authorBlock{
    Jaeseok Byun$^1$\thanks{Equal contribution} ~~ Seokhyeon Jeong$^1$\footnotemark[1] ~~ Wonjae Kim$^2$ ~~ Sanghyuk Chun$^2$\thanks{Corresponding authors} ~~ Taesup Moon$^{1,3}$\footnotemark[2]\\
\\
{$^1$Department of ECE, Seoul National University} \\
{$^2$NAVER AI Lab} \\
{$^{3}$ Department of ASRI/INMC/IPAI/AIIS, Seoul National University}
}

\newif\ifreview 
\newif\ifarxiv \newcommand{\arxiv}{\arxivtrue}
\newif\ifcamera 
\newif\ifrebuttal 

%% file: cvpr_header.tex
\ifreview \usepackage[review]{cvpr} \fi
\ifarxiv \usepackage[pagenumbers]{cvpr} \fi
\ifrebuttal \usepackage[rebuttal]{cvpr} \fi
\ifcamera \usepackage{cvpr} \fi

\input{_macros}  %

\usepackage{xr-hyper}

\makeatletter
\newcommand*{\addFileDependency}[1]{
  \typeout{(#1)}
  \@addtofilelist{#1}
  \IfFileExists{#1}{}{\typeout{No file #1.}}
}

\makeatother
\newcommand*{\myexternaldocument}[1]{
    \externaldocument{#1}
    \addFileDependency{#1.tex}
    \addFileDependency{#1.aux}
}

\definecolor{cvprblue}{rgb}{0.21,0.49,0.74}
\usepackage[pagebackref,breaklinks,colorlinks,allcolors=cvprblue]{hyperref}
\usepackage[capitalize]{cleveref}
\crefname{section}{Sec.}{Secs.}
\crefname{table}{Table}{Tables}
\crefname{figure}{Fig.}{Figs.}

\ifarxiv \crefname{appendix}{App.}{Apps.}
\else \crefname{appendix}{Suppl.}{Suppls.} \fi

%% file: _macros.tex
\usepackage{graphicx}	
\usepackage{amsmath}	
\usepackage{amssymb}	
\usepackage{booktabs}
\usepackage{times}
\usepackage{microtype}
\usepackage{epsfig}
\usepackage{caption}
\usepackage{float}
\usepackage{placeins}
\usepackage{color, colortbl}
\usepackage{stfloats}
\usepackage{enumitem}
\usepackage{tabularx}
\usepackage{xstring}
\usepackage{multirow}
\usepackage{xspace}
\usepackage{url}
\usepackage{subcaption}
\usepackage{xcolor}
\usepackage[hang,flushmargin]{footmisc}

\usepackage{xspace}
\usepackage{graphicx}
\usepackage{wrapfig}
\usepackage{booktabs}
\usepackage{multirow}
\usepackage{multicol}
\usepackage{pifont}

\usepackage{amsmath}

\usepackage{listings}
\usepackage{xcolor}
\usepackage{float}
\newcommand{\oursfull}{Reducing Task Discrepancy of Text Encoders\xspace}
\newcommand{\ours}{RTD\xspace}

\definecolor{darkergreen}{RGB}{21, 152, 56}
\definecolor{red2}{RGB}{252, 54, 65}
\newcommand{\yesmark}{\textcolor{darkergreen}{\ding{52}}}
\newcommand{\nomark}{\textcolor{red2}{\ding{56}}}

\newcommand{\purplecolor}[1]{\textcolor{violet}{\textbf{#1}}\xspace}

\ifcamera \usepackage[accsupp]{axessibility} \fi

\ifarxiv  \fi

\newcommand{\R}[1]{{%
    \textbf{%
        \ifstrequal{#1}{1}{\textcolor{red}{R#1}}{%
        \ifstrequal{#1}{2}{\textcolor{blue}{R#1}}{%
        \ifstrequal{#1}{3}{\textcolor{magenta}{R#1}}{%
        \ifstrequal{#1}{4}{\textcolor{teal}{R#1}}{%
                           \textcolor{cyan}{R#1}%
        }}}}%
    }%
}}

%% file: 00_abstract.tex
\begin{abstract}

Composed Image Retrieval (CIR) aims to retrieve a target image based on a reference image and conditioning text, enabling controllable image searches.
The mainstream Zero-Shot (ZS) CIR methods bypass the need for expensive training CIR triplets by projecting image embeddings into the text token embedding space, forming a composed query for retrieval.
However, we highlight an inherent limitation in these projection-based CIR: a task discrepancy of text encoders between the original pre-training task of the encoders (text $\leftrightarrow$ image) and the target CIR task (image + text $\leftrightarrow$ image), which potentially negatively impacts CIR performance.
To reduce such a discrepancy, a naive solution would be to train both image and text encoders with CIR triplets in a supervised manner. Instead, we introduce \oursfull (\textbf{\ours}), an efficient text-only post-hoc framework that complements projection-based CIR methods. We devise a novel target-anchored text contrastive learning designed to enhance the capability of the text encoder for CIR. 
We also propose two key enhancements: (1) a hard negative-based refined batch sampling strategy and (2) a refined concatenation scheme to further mitigate training-inference discrepancy. 
Integrating \ours into state-of-the-art projection-based methods achieves performance comparable to, or even surpassing, resource-intensive state-of-the-art synthetic CIR triplet-based approaches only with 23 minutes of additional training on 4 A100 GPUs— up to $100\times$ faster in training.
Our code will be available upon acceptance.

\end{abstract}

%% file: 01_intro.tex
\section{Introduction}
\label{sec:intro}

Composed Image Retrieval (CIR) aims at retrieving a target image that closely resembles a reference image while reflecting the changes described in a conditioning text.
Using a query composed of image and text allows users to conduct more precise and flexible searches by specifying the desired modifications to the image through text.
Supervised CIR methods \citep{baldrati2022effective, delmas2022artemis, lee2021cosmo} have been introduced to fuse information from bi-modal query, using labeled data from the target domain in the form of triplets $(I_r,T_c,I_t)$, in which $I_r$ is a reference image, $T_c$ is a conditioning text, and $I_t$ is a target image. 
However, unlike the typical web-crawled image-text datasets \citep{laion}, acquiring sufficient triplets for training needs expensive manual human annotations.
To overcome the dependency on small-scale and human-verified triplets of the target domain, several works utilize the power of recent generative models. For example, a line of studies \citep{compodiff,covr,(CASE)levy2024data,magiclens} uses text-to-image models like IP2P \citep{instructpix2pix} or large-language models (LLM) \citep{llama,gpt} to synthesize large-scale CIR triplets for training, in place of the target-domain CIR triplets. While these methods achieve strong performance, they are often impractical due to the high computational and memory requirements for utilizing generative models.

\begin{figure*}[t]
\centering
\includegraphics[width=\textwidth]{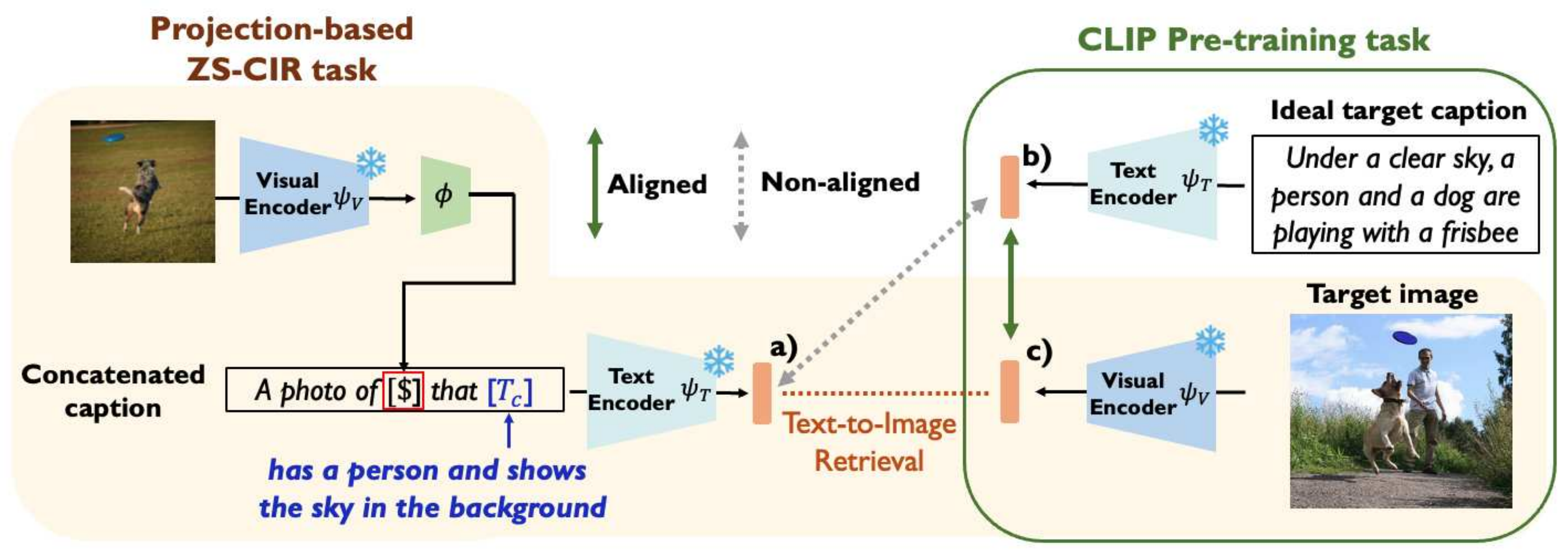} 
\vspace{-0.75cm}
\caption{ 
The task discrepancy of projection-based ZS-CIR methods between the pre-training task (image-text alignment) and the ZS-CIR task (image-text composition).
}
\label{fig:motivation_diagram}
\vspace{-0.5cm}
\end{figure*}

Another approach for removing the dependency on the CIR triplets, which often is referred to as projection-based ZS-CIR \citep{pic2word,searle,lincir,KEDs,Context-I2W,FTI4CIR}, employs an integrable projection module on top of the pre-trained, frozen, and shared VL embedding space, such as CLIP \citep{clip}.
Namely, a projection module $\phi$, which maps a CLIP image embedding to the CLIP text token embedding space, can be trained by solely using images \citep{pic2word,searle} or texts \citep{lincir}.
During inference, as illustrated in \cref{fig:motivation_diagram}, 
these methods first project the embedding of the query image to a text token embedding [\$] using the function $\phi$. This embedding is then combined with the conditioning text [$T_c$] to create the prompt ``a photo of [\$] that [$T_c$]'', which is used as a query for the text-to-image retrieval.

The core assumption of the projection-based CIR is that the pre-trained text encoder should be robust enough to combine information from both the projected text token embedding and the conditioning text. 
However, we argue that this can cause a significant \emph{task discrepancy} for the pre-trained text encoder between the original image-text alignment pre-training task (of CLIP) and the CIR task.
For example, in \cref{fig:motivation_diagram}, consider an \textit{ideal caption} that accurately describes the target image. Since the CLIP text and image encoders are learned through contrastive learning, we can expect that the target image embedding (Fig. \ref{fig:motivation_diagram}c) will align well with the embedding (Fig. \ref{fig:motivation_diagram}b) of the ideal caption. In contrast, in the projection-based CIR, the text encoder receives a concatenated caption that combines the projected token [\$] and the conditioning text, not an ideal caption. 
However, as noted by \citet{combiner}, the text encoder is not typically trained to encode complex textual modifications—such as addition, negation, or replacement—to the reference image, which are common in conditioning texts. 
As a result, there is no guarantee that the textual embedding of the concatenated caption (Fig. \ref{fig:motivation_diagram}a) closely aligns with the target image embedding (Fig. \ref{fig:motivation_diagram}c), which will undermine the retrieval performance. 

To that end, we devise a complementary post-processing approach for existing projection-based CIR methods that reduces the task discrepancy of the text encoder efficiently. Ideally, both image and text backbones would be updated using expensive CIR triplets $(I_r,T_c,I_t)$; however, we achieve this using cheap \textit{text} triplets and only update the text encoder.
These text triplets $(T_r,T_c,T_t)$ consist of a reference caption ($T_r$, \textit{e.g.}, ``a dog catching a frisbee''), a conditioning text ($T_c$, \textit{e.g.}, ``change dog to cat''), and a target caption ($T_t$, \textit{e.g.}, ``a cat catching a frisbee''), respectively. 
Given that the reference captions $T_r$ are readily available from conventional caption datasets, $T_c$ and $T_t$ can be automatically generated without any human labor \citep{cirr,fashioniq} or intensive resources \citep{compodiff,covr,(CASE)levy2024data,magiclens}. 
Using these triplets, we introduce a \emph{target-anchored text contrastive learning}, in which the text encoder updates the embedding of the concatenated caption $T_{r+c}$ of $T_r$ and $T_c$ (\textit{e.g.}, ``a dog catching a frisbee''+ ``change the dog to a cat'') to align closely with the fixed embedding of the target caption $T_t$ from the frozen text encoder.
We further enhance this text-only training approach with two key components: a batch sampling strategy that ensures hard negatives in each mini-batch and a refined concatenation scheme for $T_r$ and $T_c$ designed to reduce the training-inference task discrepancy caused by our text-only training framework.

Our experimental results demonstrate that our proposed method, dubbed as \textbf{\ours} (\oursfull), consistently and considerably improves the performance in diverse evaluation datasets (CIRR \citep{cirr}, CIRCO \citep{searle}, FashionIQ \citep{fashioniq}, COCO object composition \citep{pic2word}, and GeneCIS \citep{genecis}), when integrated with existing projection-based CIR methods (SEARLE \citep{searle}, Pic2Word \citep{pic2word}, LinCIR \citep{lincir}, Context-I2W \citep{Context-I2W}, and FTI4CIR \citep{FTI4CIR}).
Consequently, compared to resource-intensive state-of-the-art synthetic CIR triplets-based method like MagicLens \citep{magiclens}, \ours achieves comparable results, with less than a 1\% gap in CIRR R@1 and 0.1\% in FashionIQ R@10 when combined with FTI4CIR (ViT-L/14). When using larger or fine-tuned backbones (ViT-G/14 or FSC-CLIP \citep{fsc}) with LinCIR, \ours achieves superior results than MagicLens in those metrics.

We note that \ours achieves the above results with significantly higher training efficiency in both data generation and training time. Specifically, our rule-based text triplet generation requires less than 10 minutes to construct 1M text triplets—570$\times$ faster than CIR triplet generation from Compodiff \citep{compodiff}. With 4 NVIDIA A100 GPUs, the additional training time for \ours is just 23 minutes for training the ViT-L/14 backbone, making it approximately 10 to 100$\times$ faster than synthetic CIR triplets-based methods. Even with a significantly larger backbone (ViT-G/14), training takes only 2.5 hours, remaining negligible in comparison. %
Moreover, we investigate the effectiveness of \ours across various text triplet generation strategies and different backbone sizes (ViT-B/32, ViT-L/14, ViT-G/14) and types (FSC-CLIP \citep{fsc}, CLIP \citep{clip}), showing its reproducibility and compatibility.

%% file: 02_related.tex
\section{Related Work}
\label{sec:related}

\paragraph{Composed Image Retrieval.}
Unlike supervised CIR methods \citep{combiner, baldrati2022effective, delmas2022artemis, lee2021cosmo}, projection-based CIR methods \citep{pic2word,searle,lincir,FTI4CIR,Context-I2W,KEDs,image2sentence} are built upon the frozen CLIP model, where a projection module $\phi$ is trained without CIR triplets. Each method employs a different training scheme for different training schemes for $\phi$ (See \cref{sec:ExpSetup} for details). Several approaches avoid the need for any training by employing interpolation techniques \citep{slerp}, while others leverage powerful yet resource-intensive models such as LLMs and captioners \citep{cirevl,ldre}. %
While our approach is built upon projection-based CIR methods, its training strategy is closely related to another category of CIR methods \citep{compodiff,covr,(CASE)levy2024data,magiclens}
that use synthetically generated expensive CIR triplets.
However, our method stands out by utilizing text triplets and updating only the text encoder, resulting in more efficient training while maintaining strong performance.

\paragraph{Task discrepancy between the CLIP pre-training task and CIR.}
Combiner \citep{combiner} updates the text encoder to minimize the gap between the target caption feature and the sum of the reference image and conditioning text features. 
FashionERN \citep{fashionern} addresses reference image dominance by introducing a separate branch to amplify the impact of the conditioning text.
However, both Combiner and FashionERN require expensive CIR triplets $(I_r,T_c,I_t)$ for training, whereas our approach uses cheap, automatically generated text-only triplets  $(T_r,T_c,T_t)$.
As another example, \citet{mtcir} synthesize a triplet of an original image, its caption, and the masked image, treating them as the target image, conditioning text, and reference image, respectively. This approach, however, still has a gap between conditioning text (\eg, ``change dog to cat'') and image caption (\eg, ``a dog catching a frisbee''); furthermore, it needs the full fine-tuning of the CLIP model, resulting in changing the visual embeddings in the retrieval database. In contrast, \ours directly uses the conditioning texts for training and does not change the target visual encoder, enabling the reuse of pre-extracted CLIP embeddings.
Lastly, CIReVL \citep{cirevl} or LDRE \citep{ldre} reduce task discrepancy by generating descriptive captions of the composed query using a large captioning model \citep{blip2} and LLM \citep{gpt}. While effective without any training, they rely on expensive inference steps and require carefully crafted prompts by skilled users. Whereas, \ours is fully automated, human-free, and more efficient during inference.

%% file: 03_method.tex
\section{Main Method}
\label{sec:method}

\subsection{Obtaining text triplets}
\label{subsec:text generation}

To address the task discrepancy,
we collect cheaply and automatically generated text triplets $(T_r, T_c, T_t)$, instead of directly using the expensive CIR triplets $(I_r, T_c, I_t)$. 
Given reference caption $T_r$ from conventional caption datasets, conditioning text $T_c$ and target text $T_t$ can be generated using two strategies: via large language models (LLMs) \citep{compodiff,instructpix2pix,(CASE)levy2024data,covr} or, more efficiently, through rule-based templates \citep{compodiff}. 
We investigate both strategies and demonstrate that \ours consistently improves performance across them.

For the LLM-based strategy, we use the publicly available text triplets by CompoDiff \citep{compodiff}, which are employed in our main experiments if not specified. These triplets are generated by taking a given caption $T_r$ as an input of the fine-tuned LLM, whose output predicts the corresponding conditioning text $T_c$ and the target caption $T_t$. 
Previous works, such as IP2P \citep{instructpix2pix}, CoVR \citep{covr}, and CASE \citep{(CASE)levy2024data}, have also explored generating text triplets using LLMs, differing in LLM model types, input data, and fine-tuning strategies. 
The original purpose of these text triplet generations is to construct CIR triplets $(I_r, T_c, I_t)$, but all these works also release the corresponding text triplets used for their CIR triplet construction.
In addition to these publicly available text triplets, we also implement and evaluate an efficient in-context learning generation strategy using LLaMA3-8B \citep{llama3} without additional fine-tuning. 
We conduct experiments with all the aforementioned text triplets in \cref{tab:text_generation} and observe that \ours consistently delivers significant enhancements.

For cheap, rule-based strategy, 
we can extract a keyword (\eg, ``dog'') from $T_r$ (``a dog catching a frisbee'') and replace it with a randomly chosen keyword (\eg, ``cat'') \citep{compodiff}, forming $T_c$. The conditioning text is then generated automatically by using pre-defined templates (\eg, ``change \texttt{[original keyword]} to \texttt{[altered keyword]}'').
Our experiments show that this simple rule-based variant performs similarly to the LLM-based one.

More detailed explanations and examples of each strategy are provided in \cref{subsec:appedix_triplet_details}, and a comprehensive comparison of generation costs is provided in  \cref{sec:appendix_training_efficiency}.

\begin{figure*}[t]
\centering
\includegraphics[width=\textwidth]{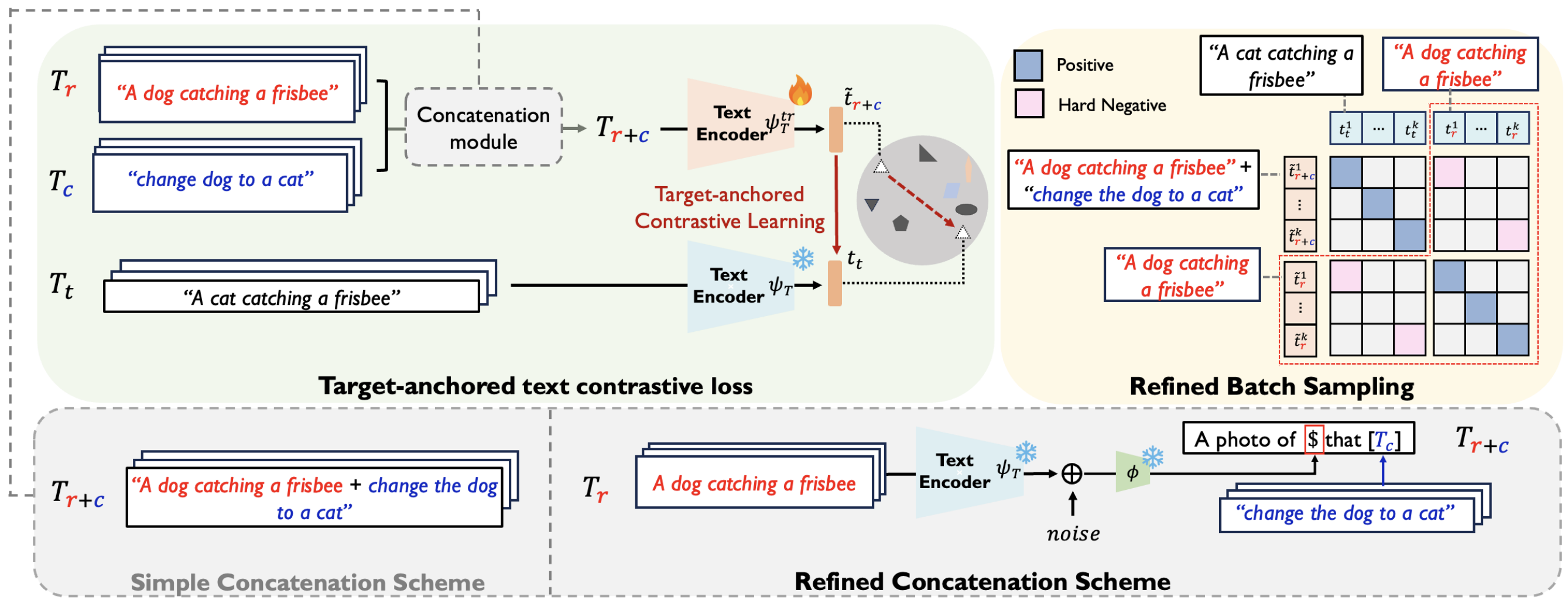} 
\vspace{-0.75cm}
\caption{Overview of \ours.}
\label{fig:method_diagram}
\vspace{-1em}
\end{figure*}

\subsection{Target-anchored text contrastive learning}
\label{sec:target_anchored}

Now, we explain our approach to update the text encoder for mitigating the task discrepancy solely with the generated text triplets $(T_r,T_c,T_t)$.
We first assume that there exists a pre-trained projection module $\phi$ obtained by the projection-based ZS-CIR methods \citep{pic2word,searle,lincir}. Recall that for a given reference image $I_r$ and conditioning text $T_c$, the final composed feature is generated by passing the text prompt ``a photo of $\phi(\psi_V (I_r))$ that $T_c$'' to the text encoder $\psi_T$, where $\psi_V$ is the visual encoder and $\phi$ is the projection module (See \cref{fig:motivation_diagram}). We aim to update the text encoder $\psi_T$ to reduce the discrepancy between the pretext task and ZS-CIR task using the text triplets
while maintaining $\psi_V$ and $\phi$ frozen.

\noindent\textbf{Target-anchored text contrastive loss.}
We apply contrastive learning using a paired caption $(T_{r+c},T_t)$, where $T_{r+c}$ denotes a concatenated caption of reference caption $T_r$ and conditioning caption $T_c$. Namely, we let the representation of the concatenated caption closely approximate that of the target caption.
However, solely updating the text encoder while fixing the image encoder can break the alignment between image and text encoders. To prevent the issue, we extract the text embedding of $T_t$ using the frozen text encoder $\psi_T$, while the concatenated caption $T_{r+c}$ is extracted from the learnable text encoder $\psi_T^{tr}$, initialized from $\psi_T$. Here, we assume that as the target caption $T_t$ is a standard caption, a text embedding $\psi_T(T_t)$, is well-aligned with the frozen image embedding space.
Following the assumption, we fix the target textual embedding as an anchor point to maintain the pre-trained alignment while learning new relationships.
As shown in \cref{subsec:ablation_studies}, this anchoring is essential for fine-tuning the text encoder with our objective. 

Now, we introduce our target-anchored text contrastive loss $\mathcal{L}_{TCL}$ using two text encoders: a frozen pre-trained text encoder $\psi_T$ and a learnable text encoder $\psi_T^{tr}$ which is initialized with $\psi_T$.
Textual latent embeddings $\tilde{t}_{r+c}$ and $t_t$ are extracted from $\psi_T^{tr}$ and $\psi_T$, respectively. 
Namely, $\tilde{t}_{r+c}= \psi_T^{tr}(E_w^{tr}(T_{r+c}))$ and $t_{t}= \psi_T(E_w(T_{t}))$, where $E_w$ is a word embedding layer.  
We aim to tune $\psi_T^{tr}$ to minimize the distance between the concatenated textual embedding $\tilde{t}_{r+c}$ and the target textual embedding $t_t$ while maximizing the distance from other textual embeddings within the batch.
We employ a symmetric InfoNCE loss \citep{simclr, palavra}, as follows:

{\small
\begin{align} \label{eq:loss_ct}
    \mathcal{L}_{TCL} =  &\frac{1}{B}  \sum^{B}_{k=1} -\log\Bigg({\frac{e^{( c(\tilde{t}_{r+c}^k, t_t^k) / \tau )}}{\sum\limits^B_{j=1} {e^{( c(\tilde{t}_{r+c}^k, t_t^j) / \tau )}}+{\sum\limits_{j\neq k} {e^{( c(t_t^k, t_t^j) / \tau )}}} }}\Bigg) \nonumber \\ 
    &- \log\Bigg({\frac{e^{( c(t_t^k, \tilde{t}_{r+c}^k) / \tau )}}{\sum\limits^B_{j=1} {e^{( c(t_t^k, \tilde{t}_{r+c}^j) / \tau )}} +  {\sum\limits_{j\neq k} {e^{( c(\tilde{t}_{r+c}^k, \tilde{t}_{r+c}^j) / \tau )}}} }} \Bigg)
\end{align}
}
in which $c(\cdot,\cdot)$ denotes the cosine similarity, $B$ is the batch size, and $\tau$ is a temperature. %

\noindent\textbf{Refined batch sampling strategy for hard negatives.} To further enhance the efficacy of updating the text encoder, we devise a simple yet effective batch sampling strategy that incorporates pairs of $(T_{r+c},T_t)$ and $(T_r,T_r)$ within the same batch. For example, as presented in \cref{fig:method_diagram}, a pair such as ($T_{r+c}$: ``A dog catching a frisbee + change dog to cat'', $T_t$: ``A cat catching a frisbee'')  is sampled along with its corresponding reference pair ($T_{r}$: ``A dog catching a frisbee'', $T_r$: ``A dog catching a frisbee'') in the same batch. This setup ensures that the concatenated text $T_{r+c}$ and its corresponding reference text $T_r$ implicitly act as hard negatives for each other, as their semantics are much more similar ($T_{r+c}$ is derived from $T_r$) than those of other randomly sampled texts in the batch. 
Explicitly distinguishing the embedding of $T_{r+c}$ from that of $T_r$ conceptually aligns well with CIR task, as it encourages the model to better capture modifications (in $T_c$).
Moreover, we believe including $(T_r,T_r)$ pairs in the contrastive learning helps the learnable text encoder $\psi_T^{tr}$ remain closely aligned with the pre-trained encoder $\psi_T$.

\noindent\textbf{Refined concatenation of reference and conditioning texts.}
As we use ``a photo of [\$] that [$T_c$]'' for inference, a naive concatenation strategy also can suffer from training-inference task discrepancy.
To tackle this issue, instead of simply concatenating the $T_r$ and $T_c$, we use the prompt ``a photo of [\$] that [$T_c$]'' for updating the text encoder, where [\$] is obtained by the reference caption $T_r$ with the projection module $\phi$. 
Instead of obtaining a pseudo-word token with latent image embedding $v$, we utilize a textual latent embedding from the reference caption $T_r$, \ie, $\phi(t_r)$.
However, \citet{lincir} showed that naively replacing the image encoder with the text encoder for the input of $\phi$ will suffer from the modality gap \citep{modality_gap},
a phenomenon where text and image embeddings have a gap between them.
We tackle this issue by injecting random noise into the textual token representation before it is processed by $\phi$, following \citet{lincir}.
More analyses on variations of noise are in \cref{subsec:appendix_noise_injection}.

\cref{fig:method_diagram} illustrates the overview of \ours.
We use CLIP backbone and pre-trained projection module $\phi$ produced by the existing projection-based CIR methods. The text encoder is trained using the proposed
loss function (\cref{eq:loss_ct}) while applying the refined batch sampling and concatenation scheme.
During inference, the procedure mirrors that of existing projection-based CIR methods, except we utilize the updated text encoder $\psi_T^{tr}$ instead of the frozen one $\psi_T$. 
Note that our method only updates the text encoder while the image encoder and the projection module are frozen.

\noindent\textit{Remark 1: Low data acquisition cost.} 
The text triplets we use can be obtained at a significantly lower cost than CIR triplets. First, the text triplet generation process requires only an easily obtainable caption dataset for generation, whereas other approaches that generate CIR triplets \citep{covr,(CASE)levy2024data,magiclens} necessitate image or video datasets along with an additional collection phase for semantically similar images or videos.
Second, text triplet generation can avoid resource-intensive text-to-image generation \citep{compodiff, instructpix2pix}, making it 15$\times$ faster than the CIR triplet generation process used in CompoDiff \citep{compodiff}.
Furthermore, if we choose the rule-based text triplet generation strategy, the process becomes 570$\times$ faster than the full CIR triplet generation with images; 1M text triplets generation takes just 0.1 hours.
Lastly, the storage for the generated text triplets takes up only 100MB, whereas storing a similar quantity of images requires significantly more space (\eg, 400GB for CC3M \citep{cc3m}). More details can be found in \cref{sec:appendix_training_efficiency}.

\begin{table}[h!]
    \centering
    \vspace{-0.15cm}
    \caption{\textbf{Training and inference time comparisons.}
    \textcolor{violet}{Purple} denotes the additional resource required for \ours. Inference time is measured on a single A100 GPU for a single image. Note that the training hours for SEARLE, LinCIR for 8 A100 GPUs \citep{lincir} and their RTD variants are trained on 4 A100 GPUs. CompoDiff is trained on 128 A100 and MagicLens is trained on 128 TPUs.
    }
    \vspace{-0.25cm}
    \resizebox{\columnwidth}{!}{
    \input{Table/time_comparison}

    }
    \label{tab:time_comparison}
  \vspace{-0.35cm}
\end{table}

\noindent\textit{Remark 2: Training and inference efficiency.}
As reported in \cref{tab:time_comparison}, the \textit{additional} training cost of \ours is small—0.38 hours for ViT-L/14 on 4 A100 GPUs. In contrast, CompoDiff and MagicLens require 231 hours on 128 A100 GPUs and 6 hours on 128 TPUs, respectively, making the cost of \ours negligible. Even with a larger ViT-G/14 backbone, training \ours takes only 2.5 hours on 4 A100 GPUs—still faster than synthetic CIR triplet-based methods and even comparable to SEARLE in ViT-L/14 (4.2h).
This training efficiency mainly stems from the advantage of text-only training.
As highlighted by \citet{lincir}, the training complexity for the text encoder is remarkably lower than that for the visual encoder due to the relatively short token lengths of texts ($\approx$12) compared to images (256). The average inference time of the CLIP ViT-L/14 image encoder is $\times$ 3.5 times slower than that of the text encoder. 
Another source of efficiency is that RTD requires relatively few iterations (approximately 2000), as the text encoders are already pre-trained and only require minor adjustments to learn the relationships between text triplets.
Moreover, in \cref{subsec:more_efficient_variants}, we present a more efficient implementation by selectively updating only a few layers of the text encoder. 
Along with efficient training, \ours retains the fast inference speed of the original projection-based methods, remaining 150$\times$ faster than training-free CIReVL.

%% file: Table/time_comparison.tex
\begin{tabular}{l c c c c}
        \toprule
        Method & CLIP Backbone  & Training (h) & Inference (s)\\
        \midrule
        SEARLE & L/14  & 4.2h & 0.02s \\
        \rowcolor{gray!20} SEARLE \color{violet}(+ RTD) & L/14  & 4.58h \color{violet} (+0.38h) & 0.02s \\
        LinCIR & L/14   & 0.5h & 0.02s \\
        \rowcolor{gray!20} LinCIR \color{violet}(+ RTD) & L/14 & 0.88h \color{violet} (+0.38h) & 0.02s \\
        LinCIR &G/14   & 0.8h & 0.05s \\
        \rowcolor{gray!20} LinCIR \color{violet}(+ RTD) &G/14  & 3.3h \color{violet} (+2.5h)& 0.05s \\ \hline
        CompoDiff & L/14  & 231h & 0.12s \\
        MagicLens & L/14   & 6h & unknown \\ \hline
        CIReVL & L/14  & - & 3.02s \\
        \bottomrule
\end{tabular}

%% file: 04_experiment.tex
\section{Experiments}
\label{sec:Exp}

\subsection{Experimental setup}
\label{sec:ExpSetup}
\textbf{Implementation details.}
We use the AdamW optimizer \citep{adamw} with a weight decay of 0.01. The learning rate is set to $10^{-5}$, with a batch size of 512.
We select the text encoder model with the best zero-shot CIRR \citep{cirr} dev R@1 score for evaluating \ours.
We mainly use the visual and textual encoders of the CLIP ViT-L/14 \citep{clip} as our backbone.
Unless otherwise noted, we use LLM-based 2.5M text triplets provided by CompoDiff \citep{compodiff} for the training. 
We set the $\tau$ as $0.07$ in \cref{eq:loss_ct} and scale the standard deviation of Gaussian distribution as $0.5$ for the noise injection. 
More results on various noise distributions can be found in the \cref{subsec:appendix_noise_injection}. 
All experiments were conducted using four NVIDIA A100 GPUs with Python 3.8 and Pytorch \citep{pytorch}.

\noindent\textbf{Evaluation datasets and metrics.}
We compare projection-based CIR methods on five benchmark datasets: CIRR \citep{cirr}, CIRCO \citep{searle}, FashionIQ \citep{fashioniq} (FIQ), COCO object composition \citep{pic2word}, and GeneCIS \citep{genecis}. Details of each dataset are in the \cref{subsec:appendix_dataset_details}.
For CIRR, FIQ, COCO, and GeneCIS, we have reported their recall scores at the top K retrieval results (R@K). 
Since the CIRCO dataset includes multiple positive images for each query, we use a ranking-based metric—mean Average Precision scores at the top K results (mAP@K) \citep{metriclearningrealitycheck,eccvcaption}.
For the main results, we compare the results on the FIQ validation split, as well as the test sets of CIRR and CIRCO. 
 For the ablation studies and analyses, the validation splits of these three datasets are utilized. 
 GeneCIS and COCO object composition results and their detailed explanations can be found in the \cref{subsec:appendix_genecis_and_coco}.

\noindent\textbf{Baselines.} We evaluate the effect of our method when combined with publicly available projection-based ZS-CIR methods: Pic2Word \citep{pic2word}, SEARLE \citep{searle}, LinCIR \citep{lincir}, Context-I2W \citep{Context-I2W}, and FTI4CIR \citep{FTI4CIR}.
All these methods share the similar core concept shown in \cref{fig:motivation_diagram}, but use different training schemes.
Pic2Word\citep{pic2word} optimizes contrastive loss between the image embedding and its projected text embedding of ``a photo of [\$]'' to obtain the projection module $\phi$.
Similarly, SEARLE \citep{searle} employs a two-stage approach, starting with an optimization-based textual inversion phase followed by a distillation phase for the projection module $\phi$.
LinCIR \citep{lincir} introduces a language-only self-supervised task involving keyword token replacement by letting the original text embedding and the replaced text embedding whose keyword tokens are changed to the projected original text embedding by $\phi$.
Context-I2W \citep{Context-I2W} refine the projection module by selecting relevant visual information  \citep{Context-I2W}. Lastly, FTI4CIR \citep{FTI4CIR} separately maps images into subject- and attribute-oriented pseudo-word tokens \citep{FTI4CIR}. Details on how \ours is combined with them can be found in \cref{subsec:appendix_fti4cir_context_i2w}.

\begin{table*}[ht]
    \centering
    \caption{ \textbf{Comparison with other baselines.} %
    Note that this comparison is not entirely fair due to differences in backbone models and training data. {\color{violet} Purple} denotes the performance gain from our method, while {\color{red} red} and {\color{blue} blue} highlight the best and second-best scores, respectively.
    }
    \vspace{-0.25cm}
    \resizebox{\textwidth}{!}{
    \input{Table/rebuttal_main_table}

    }
    \label{tab:appendix_comparison}
    \vspace{-0.45cm}
\end{table*}

We train these methods with the CLIP ViT-L/14 in our main experiments. 
To further verify the compatibility of \ours, we also report results for the ViT-B/32 backbone with SEARLE, Pic2Word, and LinCIR. 
Moreover, we conduct experiments with a larger backbone (ViT-G/14) and a fine-tuned CLIP model for enhanced compositionality (FSC-CLIP \citep{fsc}) for LinCIR, enabled by its fast training capability.
We use the publicly available pre-trained model for SEARLE (ViT-B/32, ViT-L/14), Pic2Word (ViT-L/14), Context-I2W (ViT-L/14), and FTI4CIR (ViT-L/14). Otherwise, we reproduce the results using the official implementation.
When reproducing, we adhere to the same settings in the original papers. For example, we select the final last epoch model for the Pic2Word ViT-B/32 model and choose the model based on the best zero-shot CIRR dev R@1 score for LinCIR. 
We additionally compare our method with a diverse set of CIR approaches, including another projection-based ZS-CIR method (KEDs \citep{KEDs}), another attempt to address task discrepancy (MT-CIR \citep{mtcir}), approaches leveraging synthetically generated CIR triplets (CoVR \citep{covr}, CASE \citep{(CASE)levy2024data}, MagicLens \citep{magiclens} and Compodiff \citep{compodiff}), and the training-free methods (CIReVL \citep{cirevl}, LDRE \citep{ldre}).

\subsection{Main results}
\cref{tab:appendix_comparison} shows the overview of comparison results with state-of-the-art CIR methods.
First, we assess the impact of integrating \ours with existing projection-based CIR methods (SEARLE, Pic2Word, LinCIR, Context-I2W, and FTI4CIR) across CIRR, CIRCO, and FashionIQ. \ours consistently boosts performance, yielding an average improvement of over 2.78 points.
This trend also holds for the GeneCIS and COCO object composition datasets, as detailed in \cref{subsec:appendix_genecis_and_coco}.

Second, we compare the integration of \ours with those leveraging synthetically generated CIR triplets (CoVR, CASE, MagicLens, and Compodiff). We focus on this comparison to ensure fairness, as leveraging text triplets and updating the textual backbone of RTD may not be fully aligned with the standard projection-based CIR setting.
We observe that \ours delivers competitive or superior performance compared to these resource-intensive synthetic CIR triplet-based methods, while being significantly more efficient—over 10–100$\times$ faster, as noted in \textit{Remark-2} of \cref{sec:method}.
For example, with the same CLIP ViT-L/14 backbone, LinCIR + \ours outperforms MagicLens and CoVR in FashionIQ R@10 and R@50 while FTI4CIR + \ours or Context-I2W + \ours achieve performance on par with or exceeding that of  MagicLens, MT-CIR, and Compodiff in CIRR metrics.
When combined with larger (ViT-G/14) or fine-tuned (FSC-CLIP) backbones, \ours achieves the best or second-best result on the CIRR and FashionIQ benchmarks.
We believe this flexibility of using different backbones 
underscores the advantage of efficient text-only training of \ours, inherited by LinCIR \citep{lincir}.
We reemphasize that even with a larger backbone like ViT-G/14, \ours remains significantly faster in training than existing synthetic triplet-based methods, as highlighted in \cref{tab:time_comparison}.

Lastly, across both backbones (ViT-L/14, G/14), LinCIR + \ours mostly outperforms training-free CIReVL and LDRE, except on CIRCO metrics. As shown in \cref{tab:time_comparison}, CIReVL is significantly slower (150$\times$) than projection-based methods (including \ours), while LDRE incurs even greater inference time due to its ensemble-based strategy.

\begin{table}[htbp!]
    \vspace{-0.3cm}
    \centering
    \caption{\textbf{Ablation study.}  Unlike in \cref{tab:appendix_comparison}, for ablation studies and analyses, validation splits of three CIR datasets are used for evaluation. We measure the impact of TCL loss (\cref{eq:loss_ct}), refined batch sampling (RB), and refined concatenation scheme (RC). All models are based on LinCIR ViT-L/14. The first row denotes the vanilla LinCIR without \ours. ``Avg'' denotes the average of all reported metrics. \textbf{Bold} indicates the best result.
    }
    \vspace{-0.3cm}
    \resizebox{\columnwidth}{!}{
    \LARGE

\input{Table/main_ablation}

    }
    \label{tab:ablation_method}
        \vspace{-0.5cm}

\end{table}
\subsection{Ablation studies} \label{subsec:ablation_studies}
\cref{tab:ablation_method} presents the effectiveness of the proposed components: target-anchored text contrastive loss (TCL), refined batch sampling (RB), and refined concatenation scheme (RC). 
All evaluation results are on the validation splits.
All model variants use ViT-L/14 and a projection module $\phi$ from LinCIR, making the results in row 1 indicative of the original performance of LinCIR.
We first compare the impact of the text pairs fed into TCL loss. We compare our design choice $(T_{r+c},T_t)$ (from the generated text triplets) with $(T_r,T_r)$, which is the sole option for constructing a pair given a single conventional caption $T_r$.
The results demonstrate that, on average, using generated triplets (3rd row) is more effective than using original conventional text pairs (2nd row), particularly in the CIRR and CIRCO datasets.
In addition, RB (4th row) and RC (6th row) significantly enhance the overall performance, demonstrating the effectiveness of these components.
Finally, we measure the impact of using the frozen text encoder for target caption $T_t$, denoted as ``Anchor'' in the table. 
Significant performance degradation is observed when the learnable text encoder is used for extracting the embedding of the target caption $T_t$ (5th row) compared to the target-anchored case (4th row), supporting the importance of the anchoring design choice. 

\begin{table}[htbp]
\vspace{-0.5em}
    \centering
    \caption{T2I retrieval performance of different text encoders on CIRCO val set.  ``Update (pair)'' refers to the setting in the second row of \cref{tab:ablation_method}, which uses contrastive learning with the pair $(T_r, T_r)$. 
    \textbf{Bold} indicates the best result, excluding the first row (oracle case).
    }
    \vspace{-1em}
    \resizebox{0.8\linewidth}{!}{
    \input{Table/motivation_retrieval}
    }
    \label{tab:motivationretrieval}
    \vspace{-1.5em}
\end{table}

\subsection{Anaylses on our core motivation} \label{subsec:analyses_motivation}

We conduct experiments to validate our core motivation (reducing task discrepancy) and verify that the observed gains stem from it. Other details follow \cref{subsec:ablation_studies}.

\noindent\textbf{Can \ours really reduce the task discrepancy of the text encoder?}
We first quantitatively verify whether \ours \textit{indeed} reduces the task discrepancy. We first conduct a controlled experiment that measures the text-to-image (T2I) retrieval performance of the text encoder with conditional texts.
We retrieve the target images $I_t$ with the concatenated text query $T_{r+c}$ or the ideal target caption $T_t$. 
If our text encoder successfully handles the discrepancy due to the concatenated caption, the text encoder updated by \ours will perform better than the frozen one or ``Update (pair)'', which is updated by contrastive learning with the $(T_r, T_r)$ pair (corresponding to row 2 of \cref{tab:ablation_method}).
We use the CLIP ViT-L/14 and CIRCO \citep{searle} validation dataset for evaluation.
Since the CIRCO dataset only has CIR triplets $(I_r,T_c,I_t)$, we use the BLIP \citep{blip} captioner to generate $T_r$ and $T_t$ corresponding to the $I_r$ and $I_t$, respectively.
Here, the simple concatenation scheme is applied for the text query $T_{r+c}$ in all cases for a fair comparison.
\cref{tab:motivationretrieval} shows that when the text encoder is either frozen or updated with  $(T_r, T_r)$, the retrieval results using the concatenated caption $T_{r+c}$ are significantly worse than those using the target caption $T_t$. It supports the claim that the frozen text encoder suffers from the negative effects of task discrepancy between the pretext and CIR tasks. 
Moreover, it suggests that simply updating the text encoder with  $(T_r, T_r)$ pair fails to reduce this discrepancy.  
In contrast, the text encoder updated by \ours shows a significant improvement over the frozen text encoder or one updated with $(T_r,T_r)$, showing that it successfully reduces the task discrepancy.

We additionally measure the average cosine similarity between the composed textual features with the prompt ``a photo of $\phi(\psi_V (I_r))$ that $T_c$'' (Fig. \ref{fig:motivation_diagram}a) and the target image features (Fig. \ref{fig:motivation_diagram}c).
The similarity is measured by the LinCIR ViT-L/14 model on the CIRCO validation split.
When we use the frozen CLIP text encoder ($\psi_T$), the average similarity is 0.1. By changing the text encoder to our updated text encoder ($\psi_T^{tr}$), the similarity becomes 0.29 (+0.19).
This result shows again that \ours successfully aligns the composed query features using $\phi$ to the frozen CLIP image features.

\noindent\textbf{Is all the gain of \ours from naive text-encoder updating?}
To verify that our improvements cannot be achieved solely by updating the text encoder backbone without considering the task discrepancy, we additionally measure the results of previous methods (Pic2Word and LinCIR) when naively updating the text encoders.
Namely, after training $\phi$ while keeping all other networks frozen as in previous methods, we additionally update the text encoder using the original loss function, while fixing other modules including $\phi$. We denote this update rule as ``na\"{i}ve'' in the \cref{tab:ablation_backbone}.
Unlike \ours, we observe that just naively updating the text encoder (``na\"{i}ve'') significantly degrades the performance of the baseline. 
The results indicate that merely updating the text backbone is not beneficial for CIR; instead, mitigating task discrepancy through \ours is necessary.

\begin{table}[h!]
    \vspace{-0.3cm}
    \centering
    \caption{\textbf{Impact of the update scheme.} Two update schemes are compared: (1) using the original objective from baseline (``na\"{i}ve'') and (2) using \ours. For a fair comparison, in both schemes, $\phi$ is updated first and $\psi_T$ is updated top on the frozen modules. Other details are the same as \cref{tab:ablation_method}.
    }
    \vspace{-0.25cm}
    \resizebox{\columnwidth}{!}{

\input{Table/ablation_backbone}
    }
    \label{tab:ablation_backbone}
  \vspace{-0.45cm}
\end{table}

Next, instead of using the original objective from baselines, we update the text encoder with conventional contrastive learning using the pair $(T_r, T_r)$, which corresponds exactly to row 2 in \cref{tab:ablation_method} and row 3 in \cref{tab:motivationretrieval}. This setup can also serve as a proxy for the standalone impact of breaking the alignment of pre-trained knowledge, as noted in \citep{combiner}.
As demonstrated in \cref{subsec:ablation_studies}, the large gap (more than 2 points) between EPC (last row in \cref{tab:ablation_method}) and this proxy (row 2 in \cref{tab:ablation_method}) again shows that our gain cannot be attributed to this simple alternative (naive text encoder tuning).
\begin{figure}[h]
\centering
\vspace{-0.45cm}
\includegraphics[width=\columnwidth]{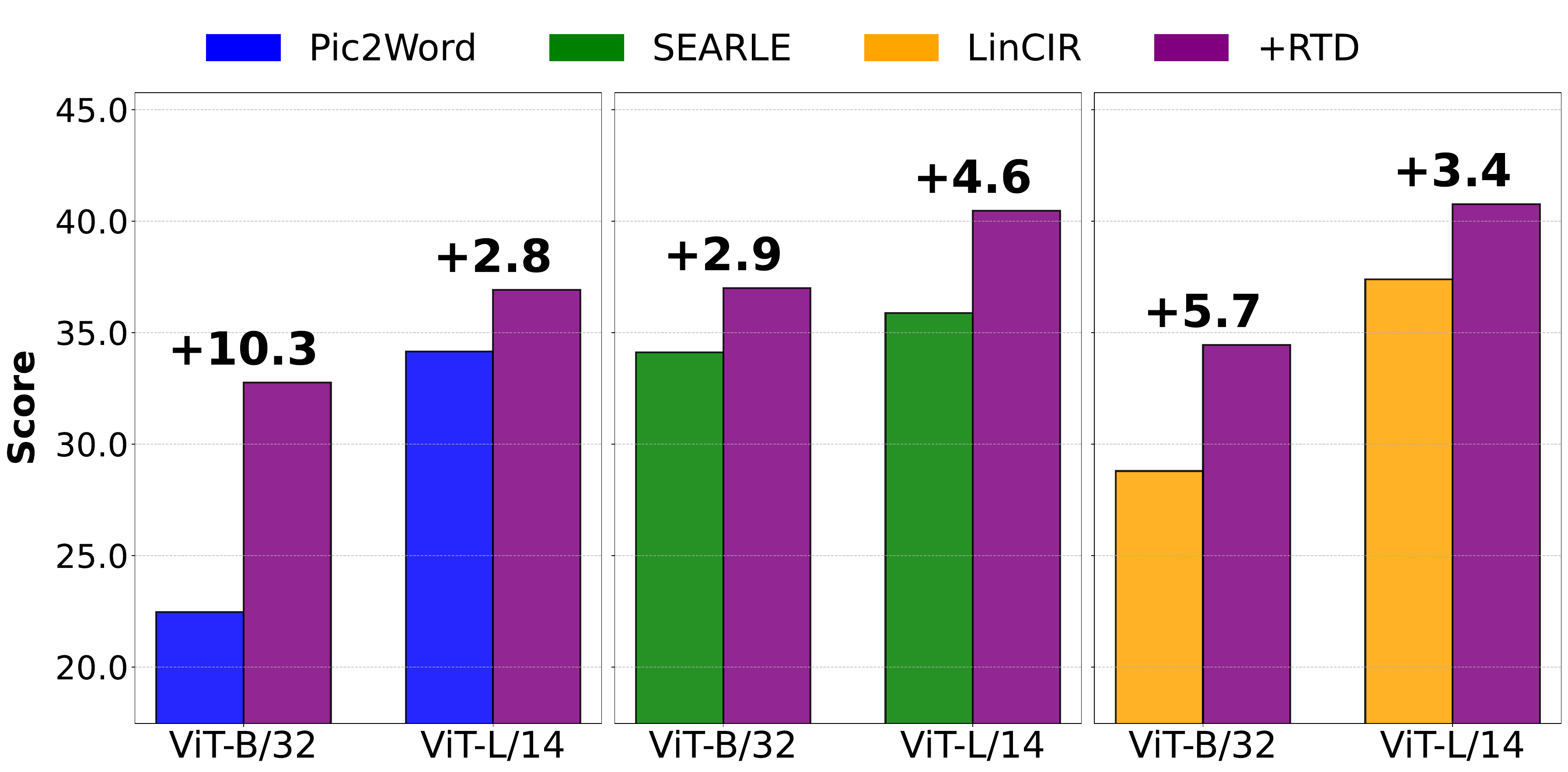} 
\vspace{-0.75cm}
\caption{ \textbf{Impact of sizes of backbone}.  The results of \ours combined with Pic2Word \citep{pic2word}, SEARLE \citep{searle}, and LinCIR \citep{lincir} across different CLIP backbones (ViT-B/32 and ViT-L/14) are shown.  Here, the score is the same metric in ``Avg'' in \cref{tab:ablation_method} and other details are the same as \cref{tab:ablation_method}. Full results are in the \cref{subsec:appendix_different_backbones}. 
}
\label{fig:comparison_chart}
\vspace{-0.25cm}
\end{figure}

\subsection{Compatibility across backbone sizes}
\label{subsec:compabibility_backbone}
In the \cref{fig:comparison_chart}, we observe that the integration of our approach with projection-based CIR methods significantly improves the performance across pioneering projection-based ZS-CIR methods (SEARLE, Pic2Word, and LinCIR) and all backbones (ViT-B/32 and ViT-L/14). For example, regardless of the choice of projection module $\phi$ and backbones, the minimum performance gain for average scores is greater than 2.8 points. The performance of \ours using the larger backbone (ViT-G/14) can be found in \cref{tab:appendix_comparison}. Details and the full results are provided in the \cref{subsec:appendix_larger_backbone}. We also provide additional qualitative retrieval results in the \cref{sec:appendix_qualitative_results}.

\begin{table}[htbp]
    \vspace{-0.2cm}
    \centering
        \caption{\textbf{The effectiveness of different types of text triplets for \ours.}
    ``In-context'' denotes an efficient implementation using in-context learning with LLaMA3-8B, without fine-tuning.
    Details and examples of each text triplet dataset are summarized in \cref{tab:langauge_example} and \cref{tab:text_generation_information}, respectively. Other details are the same as \cref{tab:ablation_method}.
}
    \vspace{-0.25cm}
    \resizebox{\columnwidth}{!}{
    \huge
    \input{Table/rebuttal_text_triplet_table}
    }
  \vspace{-0.65cm}
    \label{tab:text_generation}
\end{table}

\subsection{Impact of the text triplet generation strategies}
\label{subsec:analyses_in_ours}
As explained in \cref{subsec:text generation}, we evaluate RTD using both 1) publicly available LLM-based text triplets (from IP2P, Compodiff, CoVR, and CASE) along with efficient in-context learning-based text triplets, and 2) LLM-free, rule-based triplets. \cref{tab:text_generation} shows that RTD consistently improves CIR performance across them (+3.3 for IP2P, +3.3 for Compodiff, +3.5 for in-context learning, +4.5 for CoVR, and +0.31 for CASE, and rule-based triplets achieve 2.9, respectively). We believe this result shows the reproducibility and consistency of RTD, with the rule-based triplets performing comparably to LLM-generated ones, indicating that efficient rule-based triplets are sufficient to achieve strong CIR performance. The marginal improvement in CASE is largely due to the poor quality of text triplets resulting from its construction pipeline that prioritizes CIR triplet quality over text triplet quality, as shown in \cref{tab:langauge_example}. Further details can be found in \cref{subsec:appedix_triplet_details}, and additional analyses, including data scales related to text triplets, are provided in \cref{subsec:appendix_data_scales}.

%% file: Table/rebuttal_main_table.tex
\begin{tabular}{c|ccccccccccc}
    \toprule
    \multirow{2}{*}{\textbf{Method}} & 
    \multirow{2}{*}{\textbf{Backbone}} & \multicolumn{3}{c}{\textbf{CIRR}} & \multicolumn{3}{c}{\textbf{CIRCO}} & \multicolumn{2}{c}{\textbf{FashionIQ}}  \\
     \cmidrule(lr){3-5} \cmidrule(lr){6-8} \cmidrule(lr){9-10}
     & & \textbf{R@1} & \textbf{R@5} & \textbf{R@10} & \textbf{mAP@5} & \textbf{mAP@10} & \textbf{mAP@25} & \textbf{R@10} & \textbf{R@50}  \\
    \midrule
   
    Pic2Word \citep{pic2word} & CLIP ViT-L/14  &$24.2$ & $51.5$ & $64.1$ & $8.3$ & $9.1$ & $10.1$ & 25.3 & 44.9  \\
    
     +\ours  &   & $27.9$ \color{violet}($\mathbf{+3.6}$) & $56.2$ \color{violet}($\mathbf{+4.8}$) & $68.5$ \color{violet}($\mathbf{+4.4}$) & $9.1$ \color{violet}($\mathbf{+0.9}$) & $9.6$ \color{violet}($\mathbf{+0.5}$) & $10.7$ \color{violet}($\mathbf{+0.6}$) & 27.6 \color{violet}($\mathbf{+2.3}$) & 48.9 \color{violet}($\mathbf{+4.0}$) \\
    SEARLE \citep{searle} & CLIP ViT-L/14   & $24.9$ & $52.3$ & $65.7$ & $11.6$ & $12.7$ & $14.3$ & $25.0$ & $45.3$  \\
   \rowcolor{gray!20} +\ours  & & $26.6$ \color{violet}($\mathbf{+1.7}$) & $56.2$ \color{violet}($\mathbf{+3.9}$) & $69.0$ \color{violet}($\mathbf{+3.3}$) & $16.5$ \color{violet}($\mathbf{+4.9}$) & $17.9$ \color{violet}($\mathbf{+5.2}$) & $19.8$ \color{violet}($\mathbf{+5.4}$) & $29.3$ \color{violet}($\mathbf{+4.4}$) & $50.7$ \color{violet}($\mathbf{+5.4}$) \\
    Context-I2W \citep{Context-I2W} & CLIP ViT-L/14  & 25.6 & 55.4 & 68.6 & - & - & - & 27.9 & 49.1  \\
    \rowcolor{gray!20} +\ours  &   & $29.2$ \color{violet}($\mathbf{+4.6}$) & $58.4$ \color{violet}($\mathbf{+3.0}$) & $70.5$ \color{violet}($\mathbf{+1.9}$)& - & - & -& $28.1$ \color{violet}($\mathbf{+0.2}$)& $49.5$ \color{violet}($\mathbf{+0.4}$)  \\
    FTI4CIR \citep{FTI4CIR} & CLIP ViT-L/14  & 25.9 & 55.6 & 67.7 & 15.1 & 16.3 & 18.1 & 29.4 & 50.9  \\
   \rowcolor{gray!20} +\ours  &   & $29.1$ \color{violet}($\mathbf{+3.2}$) & $58.8$ \color{violet}($\mathbf{+3.2}$)   & $70.4$ \color{violet}($\mathbf{+2.7}$) & $16.2$ \color{violet}($\mathbf{+1.1}$) & 17.4 \color{violet}($\mathbf{+1.1}$) & 19.4 \color{violet}($\mathbf{+1.3}$) & $30.6$ \color{violet}($\mathbf{+1.2}$) & $51.7$ \color{violet}($\mathbf{+0.8}$)  \\
    LinCIR \citep{lincir} & CLIP ViT-L/14 & $23.8$ & $52.9$ & $66.5$ & $13.0$ & $14.1$ & $15.8$  & $27.4$ & $47.7$  \\
    \rowcolor{gray!20} +\ours  &  & $26.6$ \color{violet}($\mathbf{+2.9}$) & $56.2$ \color{violet}($\mathbf{+3.3}$) & $69.0$ \color{violet}($\mathbf{+2.5}$) & $17.1$ \color{violet}($\mathbf{+4.1}$) & $18.1$ \color{violet}($\mathbf{+4.0}$) & $20.1$ \color{violet}($\mathbf{+4.3}$)  & $30.2$ \color{violet}($\mathbf{+2.8}$) & $51.1$ \color{violet}($\mathbf{+3.4}$) \\
    LinCIR \citep{lincir} & FSC-CLIP \citep{fsc} ViT-L/14   & 33.6 & 63.1 & 74.5 & 14.2 & 15.4 & 17.0 & 28.3 & 49.5  \\
    \rowcolor{gray!20} +\ours &  & $35.9$ \color{violet}($\mathbf{+2.3}$) & \color{blue} $\mathbf{67.5}$ \color{violet}($\mathbf{+4.4}$) & $78.1$ \color{violet}($\mathbf{+3.6}$) & $18.3$ \color{violet}($\mathbf{+4.1}$) & $19.7$ \color{violet}($\mathbf{+4.3}$) & $21.3$ \color{violet}($\mathbf{+4.3}$)   & $31.2$ \color{violet}($\mathbf{+2.9}$) & $52.3$ \color{violet}($\mathbf{+2.8}$)   \\
     KEDs \citep{KEDs} & CLIP ViT-L/14  & 26.4 & 54.8 & 67.2 & - & - & - & 26.8 & 47.9  \\
     CIReVL \citep{cirevl}  & CLIP ViT-L/14  & 24.5 & 52.3 & 64.9 &18.6 &19.0 & 20.9  & 28.6 & 48.6  \\
    LDRE \citep{ldre}  & CLIP ViT-L/14  & 26.5 & 55.6 & 67.5 &23.4 &24.0 & 26.4  & 28.5 & 50.5  \\

    MT-CIR \citep{mtcir} & CLIP ViT-L/14  & 25.5 & 54.6 & 67.6 & 10.4 & 11.6 & 13.0  & 35.4 & 57.4  \\
    MagicLens \citep{magiclens} & CLIP ViT-L/14   & 30.1 & 61.7 & 74.4 & \color{blue} \textbf{29.6} & \color{blue} \textbf{30.8} & \color{blue} \textbf{33.4}  & 30.7 & 52.5  \\
    Compodiff \citep{compodiff} & CLIP ViT-L/14   & 26.7 & 55.0 & 72.6 & 12.6 &13.4 & 15.8  & 36.0 & 48.6  \\
    CoVR \citep{covr} & BLIP ViT-L/14  & \color{red} \textbf{39.3} & \color{red} \textbf{68.2} & \color{red} \textbf{78.9} & - & - & - & 27.7 & 44.6  \\
    CASE \citep{(CASE)levy2024data} & BLIP ViT-L/14  & 35.4 & 65.8 & 78.5 & - & - & - & - & -  \\
        \midrule
    LinCIR \citep{lincir}  & CLIP ViT-G/14  & 34.9 & 64.5 & 76.1 & 20.6 &21.9 & 24.1  & \color{blue}\textbf{44.5} & \color{red}\textbf{65.5}  \\
     \rowcolor{gray!20} +\ours  &  & $\color{blue} \textbf{36.3}$ \color{violet}($\mathbf{+1.4}$) & \color{blue} \textbf{67.5} \color{violet}($\mathbf{+3.0}$)  & \color{blue} \textbf{78.3} \color{violet}($\mathbf{+2.2}$) & 21.1 \color{violet}($\mathbf{+0.5}$) &22.3 \color{violet}($\mathbf{+0.4}$) & 24.5 \color{violet}($\mathbf{+0.4}$)  & \color{red}\textbf{46.2} \color{violet}($\mathbf{+1.7}$) & \color{red}\textbf{67.3} \color{violet}($\mathbf{+1.8}$)  \\
     Compodiff \citep{cirevl}  & CLIP ViT-G/14  & 34.7 & 64.3 & 75.1 & 15.3 &17.7 & 19.4 & 39.0 & 51.7  \\
     CIReVL \citep{cirevl}  & CLIP ViT-G/14  & 26.7 & 55.1 & 74.5 & 26.8 &27.6 & 30.0  & 32.2 & 52.4  \\
     LDRE \citep{ldre}  & CLIP ViT-G/14  & 36.2 & 66.4 & 77.3 & \color{red} \textbf{31.1} & \color{red}  \textbf{32.2} &\color{red}  \textbf{35.0}  & 32.5 & 55.4  \\
    \bottomrule
\end{tabular}

%% file: Table/main_ablation.tex
\begin{tabular}{cccc|cc|cc|cc|cc}
\toprule
\multicolumn{2}{c}{TCL} & \multirow{2}{*}{RB} & \multirow{2}{*}{RC} & \multicolumn{2}{c|}{CIRR (R)} & \multicolumn{2}{c|}{CIRCO (mAP)} & \multicolumn{2}{c|}{FIQ (R)} & \multirow{2}{*}{Avg} \\
Text & Anc & & & @5 & @10 & @10 & @25 & @10 & @50 \\
\midrule
- & - & \nomark & \nomark & 54.3 & 67.8 & 12.7 & 14.5 & 27.4 & 47.7 & \multicolumn{1}{c}{37.4} \\
$(T_r,T_r)$ & \yesmark & \nomark & \nomark & 56.0 & 69.7 & 13.4 & 15.2 & 28.2 & 48.8 & \multicolumn{1}{c}{38.5} \\
$(T_{r+c},T_t)$ & \yesmark & \nomark & \nomark & \textbf{58.2} & \textbf{71.5} & 14.4 & 16.0 & 26.9 & 47.9 & \multicolumn{1}{c}{39.2} \\
$(T_{r+c},T_t)$ & \yesmark & \yesmark & \nomark & \textbf{58.2} & 71.3 & 15.0 & 16.7 & 27.4 & 49.3 & \multicolumn{1}{c}{39.6} \\
$(T_{r+c},T_t)$ & \nomark &\yesmark & \nomark & 54.3 & 67.0 & 12.2 & 13.6 & 25.0 & 45.3 & \multicolumn{1}{c}{36.3} \\ \midrule
$(T_{r+c},T_t)$ & \yesmark &\yesmark & \yesmark & 57.9 & 71.1 & \textbf{16.1} & \textbf{17.8} & \textbf{30.2} & \textbf{51.1} & \multicolumn{1}{c}{\textbf{40.7}} \\
\midrule
\end{tabular}

%% file: Table/motivation_retrieval.tex
\setlength{\tabcolsep}{2pt}
\begin{tabular}{ccccc}
\toprule
Query & Text encoder & mAP@5 & mAP@10 & mAP@25 \\
\midrule
$T_t$ & Frozen & 18.96 & 19.31  & 21.05 \\ \hline
$T_{r+c}$ & Frozen & 10.12 & 10.71  & 12.34 \\
$T_{r+c}$ & Update (pair)  & 10.52 & 11.15 & 12.72 \\
 $T_{r+c}$ & \ours & \textbf{15.12} & \textbf{15.80}  & \textbf{17.77} \\
\bottomrule
\end{tabular}

%% file: Table/ablation_backbone.tex
\begin{tabular}{c|cc|cc|cc|c}
\toprule
\multirow{2}{*}{} & \multicolumn{2}{c|}{CIRR (R)} & \multicolumn{2}{c|}{CIRCO (mAP)} & \multicolumn{2}{c|}{FIQ (R)} & \multirow{2}{*}{Avg} \\
& @5 & @10 & @10 & @25 & @10 & @50 & \\
\midrule
Pic2Word & 51.4 & 64.4 & 8.8 & \textbf{10.1} & 25.3 & 44.9 & 32.2\\ \hline
\vspace{0.02mm}
+na\"{i}ve & 19.2 & 27.5 & 1.3 & 1.6 & 4.4 & 11.2 & 10.9\\

\multirow{1}{*}{+ \ours} & \textbf{56.6} & \textbf{69.8} & \textbf{8.8} & 9.8 & \textbf{27.6} & \textbf{48.9} & \textbf{36.9} \\
\midrule

LinCIR & 54.3 & 67.8 & 12.7 & 14.5 & 27.4 & 47.7 & 36.9\\ \hline
\vspace{0.02mm}
+na\"{i}ve & 52.7 & 66.8 & 11.4 & 13.0 & 26.3 & 45.9 & 35.5\\

\multirow{1}{*}{+ \ours }& \textbf{57.9} & \textbf{71.1} & \textbf{16.1} & \textbf{17.8} & \textbf{30.2} & \textbf{51.1} & \textbf{40.7} \\
\bottomrule
\end{tabular}

%% file: Table/rebuttal_text_triplet_table.tex
    \begin{tabular}{c|cc|cc|cc|cc|c}
        \toprule
        \multirow{2}{*}{\textbf{}}&\multirow{2}{*}{\textbf{Source}} &\multirow{2}{*}{\textbf{LLM}} & \multicolumn{2}{c|}{\textbf{CIRR (R)}} & \multicolumn{2}{c|}{\textbf{CIRCO (mAP)}} & \multicolumn{2}{c|}{\textbf{FIQ (R)}} & \multirow{2}{*}{\textbf{Avg}} \\
        & &  & \textbf{@5} & \textbf{@10} & \textbf{@10} & \textbf{@25} & \textbf{@10} & \textbf{@50} & \\
        \midrule
        LinCIR & - & - & 54.3 & 67.8 & 12.7 & 14.5 & 27.4 & 47.7 & 37.4 \\
        \cmidrule{1-10}
        \multirow{5}{*}{+RTD}         & Rule-based &\nomark & 56.7 & 70.3 & 15.0 & 17.0 & 30.4 & 51.9 & 40.2 \purplecolor{(+ 2.9)} \\         \cmidrule{2-10}

        & IP2P \citep{instructpix2pix} &\yesmark & 58.7 & 71.6 & 15.9 & 18.0 & 29.6 & 50.7 & 40.7 \purplecolor{(+ 3.3)} \\
        & Compodiff \citep{compodiff} &\yesmark  & 57.9 & 71.1 & 16.1 & 17.8 & 30.2 & 51.1 & 40.7 \purplecolor{(+ 3.3)} \\  
        & In-context &\yesmark & 59.3 & 71.8 & 15.8 & 17.5 & 29.7 & 51.4 & 
  40.9 \purplecolor{(+ 3.5)} \\  
        
        & CoVR \citep{covr} &\yesmark & 59.8 & 72.6 & 15.4 & 17.0 & 29.6 & 50.8 & 41.9 \purplecolor{(+ 4.5)} \\
        & CASE \citep{(CASE)levy2024data} &\yesmark & 56.3 & 69.3 & 11.1 & 12.7 & 26.6 & 47.8 & 37.7 \purplecolor{(+ 0.3)} \\
        \bottomrule
    \end{tabular}

%% file: 05_conclusion.tex
\vspace{-0.5em}
\section{Conclusion}
\label{sec:conclusion}

We presented \ours,  a novel post-processing approach that can be easily integrated into existing projection-based CIR methods, aimed at reducing task discrepancy of text encoders. 
Empirical evaluations demonstrate that \ours significantly boosts the performance of existing projection-based CIR methods across diverse datasets and model backbones, competing with or outperforming other resource-intensive CIR methods with much greater efficiency.

%% file: 12_appendix.tex
\clearpage 
\appendix
\setcounter{page}{1}
\renewcommand\thesection{\Alph{section}}
\setcounter{section}{0}
\onecolumn

{
    \centering
    \Large
    \textbf{\thetitle}\\
    \vspace{0.5em}Supplementary Material \\
    \vspace{1.0em}
}

\section{Additional Implementation Details}
\subsection{CIR Datasets}
\label{subsec:appendix_dataset_details}
FashionIQ \citep{fashioniq} is a dataset that contains fashion-related images from three main categories: Shirts, Dresses, and Toptee. It has a total of 30,134 triplets, which were created from 77,684 images. As the ground truth labels are not publicly available, we utilize the results from the validation set for our analysis and comparison.
CIRR \citep{cirr} encompasses a wider range of domains and contains images with more complex descriptions compared to FashionIQ.
It contains 36,554 triplets extracted from 21,552 images, which are sourced from the well-known NLVR2 natural language inference dataset \citep{(NLVR)suhr2018corpus}. 
As pointed out in previous works \citep{pic2word,lincir,searle}, CIRR and FashionIQ suffer from a significant number of false negatives, which can potentially lead to inaccurate retrieval evaluations \citep{searle,pic2word}.
To address this issue, CIRCO \citep{searle}, based on COCO images \citep{coco2014}, is recently introduced by providing multiple positive images for each query. This approach enables a more reliable and robust mAP metric \citep{metriclearningrealitycheck,eccvcaption}, which is essential for accurate evaluation of retrieval performance.

We additionally provide results on two more benchmark datasets, GeneCIS \citep{genecis} and COCO Object Composition introduced by \citep{pic2word}, in \cref{subsec:appendix_genecis_and_coco}.
GeneCIS \citep{genecis} is also constructed based on COCO images and the Visual Attributes in the Wild dataset \citep{pham2021learning}. 
GeneCIS introduces four task variations: (1) focus on an attribute, (2) change an attribute (3) focus on an object, and (4) change an object. These tasks explore different aspects of image retrieval and manipulation. 
For the COCO Object Composition task, we utilize 5000 images from the COCO validation dataset to evaluate object composition. Our objective is to retrieve an image that contains an object specified by a query image, along with scenes or objects described using text. The composed query is constructed by combining ``a photo of [\$], [$obj_1$], [$obj_2$] ... and [$obj_n$]'' where [$obj_i$] are text descriptions.

\subsection{Details of text triplets}
\label{subsec:appedix_triplet_details}
\begin{figure}[ht]
\centering
\includegraphics[width=\textwidth]{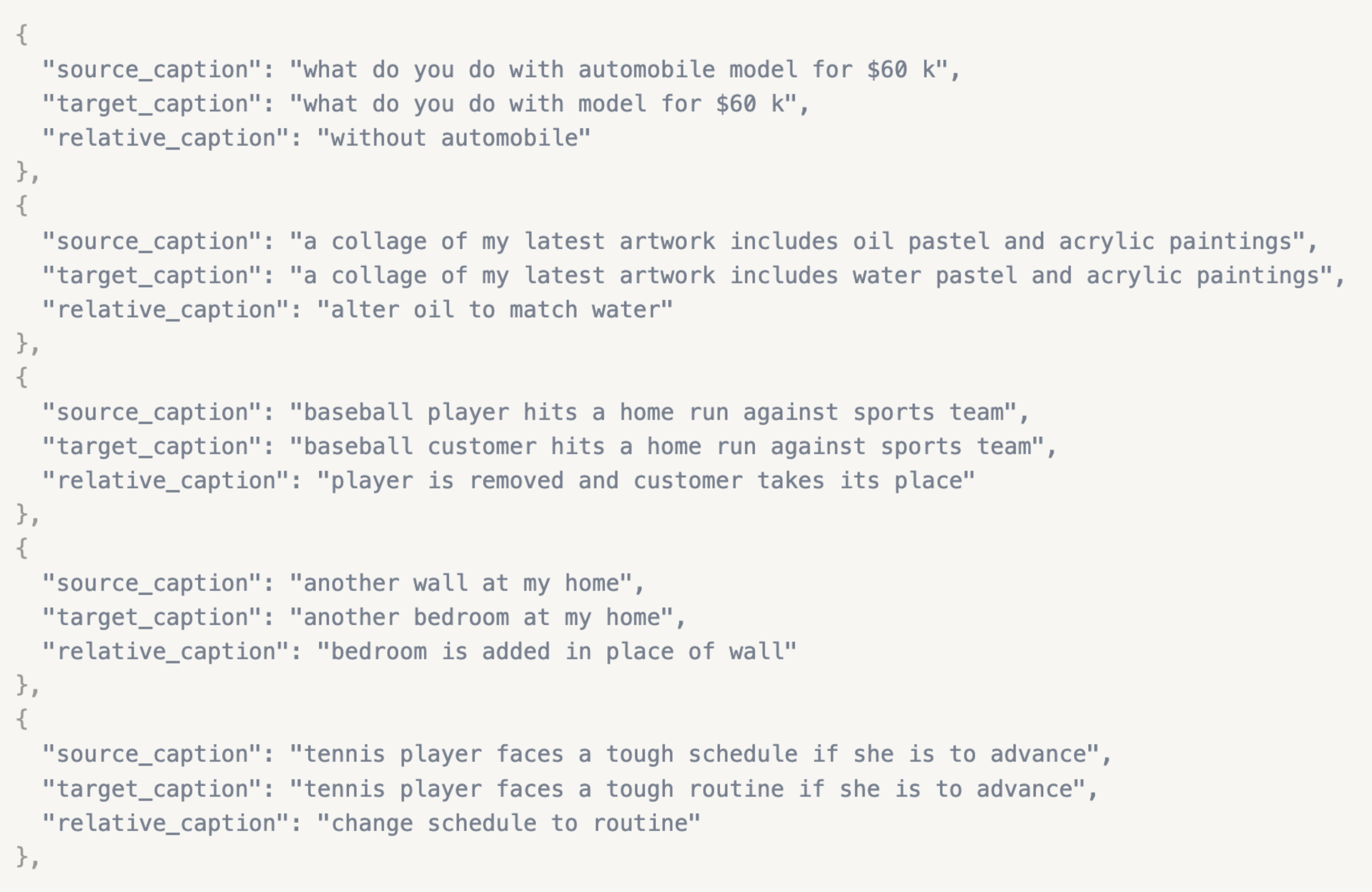} 
\caption{Example of rule-based triplet datasets}
\label{fig:template_based}
\vspace{-1em}
\end{figure}

\begin{figure}[ht]
\centering
\includegraphics[width=\textwidth]{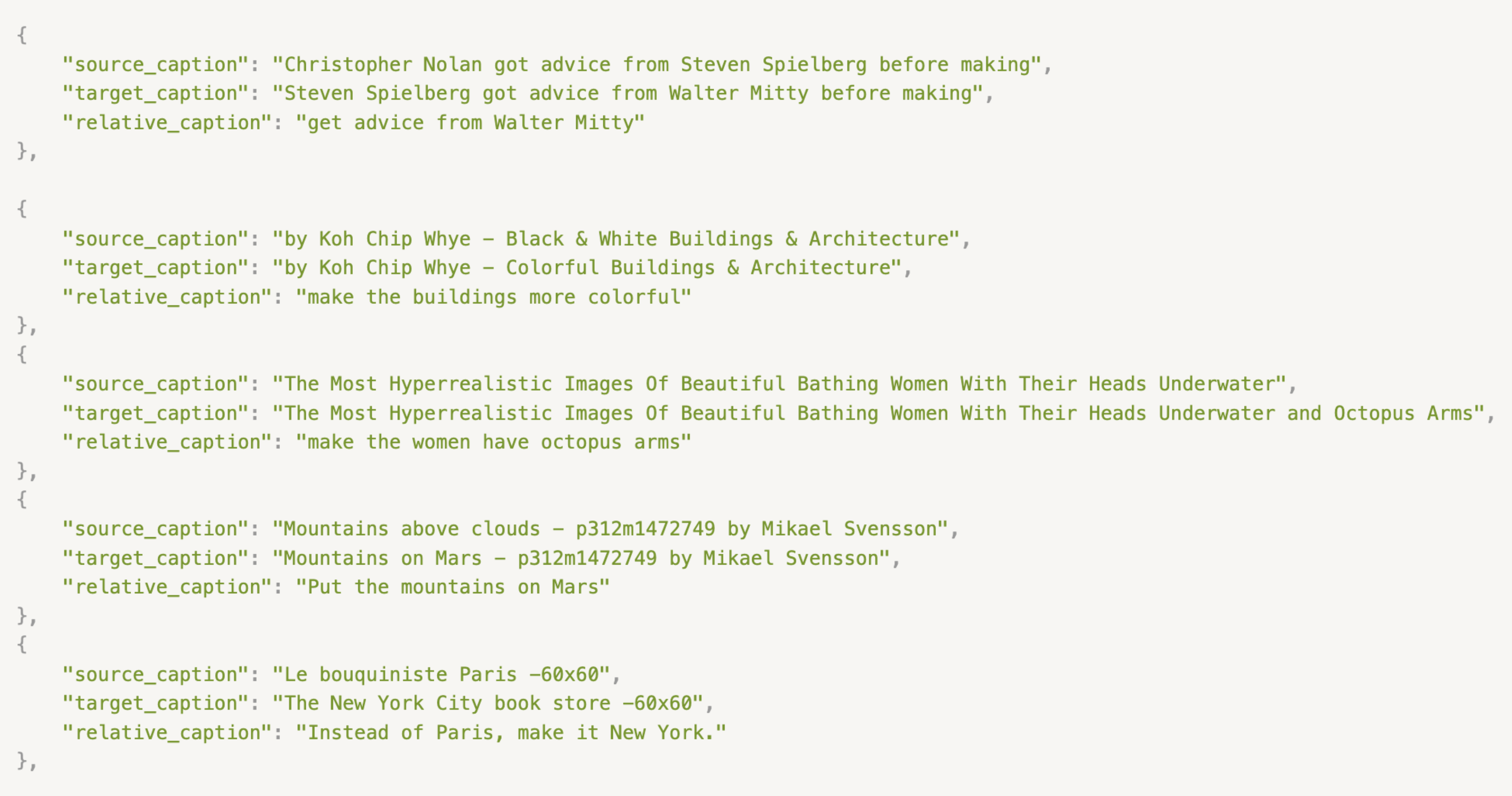} 
\caption{Example of LLM-based triplet datasets}
\label{fig:LLM_based}
\vspace{-1em}
\end{figure}

Here, we describe the details of the LLM-based and rule-based text triplet generation process. 
As shown in \cref{fig:template_based,fig:LLM_based}, which showcases examples of both LLM-based and rule-based triplets, both approaches produce natural and coherent text triplets. Note that none of the datasets used for generating text triplets overlap with the data used in the target CIR benchmarks, with the exception of the CASE dataset \citep{(CASE)levy2024data}. The source of the CASE dataset is VQA2.0 \citep{vqa2}, which is constructed from the COCO dataset \citep{coco2014}, potentially leading to overlap in cases involving COCO object composition \citep{pic2word}.

\noindent\textbf{[Detailed explanation on LLM-based triplets]} 
As described in \cref{sec:method}, besides Compodiff \citep{compodiff}, we conduct experiments using various publicly available text triplets: IP2P \citep{instructpix2pix}, COVR \citep{covr}, and CASE \citep{(CASE)levy2024data}.  Although the primary objective of these approaches is to generate CIR triplets $(I_r, T_c, I_t)$, they also produce text triplets. Below, we provide detailed descriptions of how text triplets are constructed in each approach (Note again that their final product is CIR triplets). There are two main ways to generate text triplets using LLMs: 1) generating both conditional text $T_c$ and target caption $T_t$ given reference caption $T_r$ using fine-tuned LLM for this task, such as IP2P, Compodiff; and 2) generating only conditional text $T_t$ given pairs $(T_r, T_t)$ from pre-existing captions by identifying with visually or text semantically similar such as CoVR \citep{covr}, and CASE \citep{(CASE)levy2024data}. In addition to these existing datasets, we implement an efficient in-context learning-based generation process. Examples and summaries of each dataset can be found in \cref{tab:langauge_example} and \cref{tab:text_generation_information}.

\textbf{IP2P} employs GPT-3 for text triplets generation and fine-tunes it with a human-curated small set of 700 text triplets. Namely, given reference captions $T_r$ sampled from  LAION-Aesthetics V2 6.5+ dataset \citep{LAION5B}, the corresponding conditional texts $T_c$ and corresponding target captions $T_t$ are manually written by humans. After fine-tuning on this small set of text triplets, the model generates 454k text triplets: reference captions $T_r$ from the LAION-Aesthetics V2 6.5+ dataset \citep{LAION5B} are provided as input to the fine-tuned LLM, whose output predicts the corresponding conditioning text $T_c$ and the target caption $T_t$. Note that the LAION-Aesthetics dataset is not related to the original source datasets (FashionIQ, NLVR2, and MS-COCO) used in existing CIR benchmarks (FashionIQ, CIRR, and CIRCO), ensuring no overlap with the CIR benchmarks. 

\textbf{Compodiff} enhances the scalability of the IP2P text triplet generation process by modifying the choice of LLM and expanding the fine-tuning dataset. As described in [\cite{compodiff}, Section 4], the OPT-6.7B model is utilized and fine-tuned with LoRA on the above 454k text triplets of IP2P \citep{instructpix2pix}. Then, similar to the IP2P approach, given reference captions from the LAION dataset \citep{laion}, fine-tuned LLM generates the corresponding conditioning texts and target captions. 

\textbf{COVR} starts by identifying similar caption pairs from the WebVid2M dataset \citep{webvid}, which contains 2.5 million video-caption pairs. These pairs serve as the reference captions ($T_r$) and target captions ($T_t$). Then, given these pairs ($T_r, T_t$), LLM generates conditional captions that describe the differences between the paired captions. The LLaMA-7B model \citep{llama} is utilized for this purpose and is fine-tuned on an expanded version of the above 700 manually annotated triplets used in IP2P (adding 15 annotations for more diverse cases).

\textbf{CASE} uses VQA2.0 dataset \citep{vqa2}, which consists of (image, question, answer) triplets. Given $(I, Q, A)$ triplets, complementary triplets $(I_c, Q, A_c)$ are manually selected based on visually similar image $I_c$ with three rules: 1) the premise assumed in question $Q$ holds for both $I$ and $I_c$, 2) $Q$ is logical for $I_c$, and 3) the answer $A_c$ for $I_c$ differs from $A$. Then, conditional text $T_c$ is generated by GPT-3, describing differences between image pairs $(I,I_c)$ without fine-tuning, leveraging in-context learning with a few examples. Since the VQA2.0 dataset is derived from the COCO dataset, COCO captions that match VQA2.0 images are used to form text triplets. 

As seen in \cref{tab:langauge_example}, compared to other approaches, the quality of the relationships between $T_r$, $T_c$, and $T_t$ is not always satisfactory, which results in minimal performance gain as shown in \cref{tab:text_generation}. Namely, unlike other CIR datasets that first create high-quality text triplets before generating CIR triplets, CASE generates the conditioning text $T_c$ using the reference image $I_r$ and target image $I_t$. The provided reference text $T_r$ and target text $T_t$ are taken directly from the captions of reference image $I_r$ and target image $I_t$ in the VQA2.0 dataset. Therefore, due to the poor descriptiveness of these captions and their lack of consideration for the conditioning text $T_c$, while $T_c$ can effectively explain the visual differences between $I_r$ and $I_t$, it often fails to capture the differences between $T_r$ and $T_t$ adequately.

\textbf{Efficient in-context learning} refers to our efficient implementation which uses a recent and powerful LLM, LLaMA3-8B \citep{llama3}. This approach performs in-context learning using reference captions $T_r$ from the CC3M dataset \citep{cc3m}, guided by a custom-designed prompt with a few examples of textual modifications (\eg, replace, change, remove, ...). Specifically, given a reference caption $T_r$, the prompt instructs the model to generate a target caption $T_t$, which is a complete sentence that slightly differs from the corresponding reference caption. Then, the prompt guides the model to generate conditioning text that explains the differences between $T_r$ and $T_t$, based on the above pre-defined textual modifications. Compared to Compodiff, which takes 3.8 hours to generate 1 million text triplets, this version requires only 1.5 hours. In \cref{tab:text_generation}, we verify that this more efficient version achieves competitive performance compared to the other fine-tuned LLM approaches.

\vspace{-4mm}
\begin{table}[htbp]
    \centering
    \caption{\textbf{Examples of text triplet datasets.} }
    \resizebox{\columnwidth}{!}{
    \input{Table/rebuttal_language_table_example}
    }
    \label{tab:langauge_example}
\end{table}

\begin{table}[htbp]
    \centering
    \caption{\textbf{Summaries of text triplet dataset. }}
    \resizebox{\columnwidth}{!}{

\input{Table/rebuttal_text_generation_information}

    }
    \vspace{1mm}
    \label{tab:text_generation_information}
\end{table}

\noindent\textbf{[Detailed explanation on rule-based triplets]}  
To construct rule-based triplets, we mainly follow the process described in [\cite{compodiff}, Section 4.1]. 
Firstly, given reference captions, important keywords like nouns are extracted with a part-of-speech (POS) tagger via the Spacy library. 
Then, the keyword is filtered by frequency filtering with hard thresholding to focus only on frequently occurring keywords. Specifically, we only use keywords that appear more than 100.
After applying keyword frequency filtering, the remaining keyword list is used to create caption triplets $(T_r, T_c, T_t)$. 
To generate text triplets, a keyword from the given $T_r$ is selected, and alternative keywords are extracted based on text similarity scores ranging from 0.5 to 0.7, using the SBERT all-MiniLM-L6-v2 \citep{reimers2019sentence}.
The target caption $T_t$ is then constructed by substituting the original keyword with a similar alternative. 
The conditioning text $T_c$ is generated using randomly selected pre-defined templates, as detailed in \cref{tab:template}.
Here, most of the templates are similar to that of Compodiff \citep{compodiff}. We use captions from the CC3M dataset \citep{cc3m} as reference captions $T_r$. Note that CC3M is not related to the existing CIR benchmarks, which again ensures no overlap with the CIR benchmarks.

Since the quality of the generated triplets with the above procedure may not be optimal, we employ an additional filtering process.
Compodiff \citep{compodiff} employs an additional filtering process that uses cosine similarities between generated images and texts, calculated by CLIP encoders. 
However, as we do not have images for captions, we filter the inappropriate texts using only textual information inspired by LinCIR \citep{lincir}.
Namely, we calculate the similarity between the CLIP text embedding of $T_t$ and the CLIP text embedding of ``a photo of [$\$$]'' where [\$] is obtained by $T_t$ projected by $\phi$ from LinCIR (ViT-L/14). Following LinCIR noise ($\text{Unif}(0, 1) \times \mathcal N(0, 1)$) is injected before passing through $\phi$. 
After calculating the above similarity, texts whose similarities are less than the threshold (0.75) are removed.
The same process is also applied to the reference caption $T_r$ and the intersection of filtering processes for $T_t$ and $T_r$ is used for the final dataset whose size becomes 1.3M. 
As described in \cref{subsec:appendix_filtering_experiment}, we verify that this filtering process is effective. However, this does not imply that the effectiveness of rule-based text triplets is solely dependent on the use of a projection module in the filtering process; even without filtering, the enhancement from \ours remains significant.

\vspace{3cm}

\begin{table}[ht]
    \small
    \centering
    \caption{The full 50 keyword converting templates}
    \resizebox{\columnwidth}{!}{
    \input{Table/template}

    }
    \label{tab:template}
\end{table}

\subsection{Details on integration with FTI4CIR \citep{FTI4CIR} and Context-I2W \citep{Context-I2W}.}\label{subsec:appendix_fti4cir_context_i2w} 
\textbf{FTI4CIR} \citep{FTI4CIR} enhances the fine-grained capability of the projection module $\phi$ by separately handling subjects and attributes using distinct projection modules, $\phi_{s}$ and $\phi_{a}$. To achieve this, they leverage BLIP-generated captions that explicitly separate subject and attribute information in the text domain, formatted as ``[primary subject(s)] + [detailed description].''
Unlike the subject module $\phi_{s}$, which focuses on global subject information, $\phi_{a}$ processes localized attribute details. FTI4CIR extracts attribute features by adaptively aggregating patch features via an additional transformer. However, this approach requires an extra forward pass of the transformer with full sequences of visual embeddings for $\phi_{a}$, rather than a single pooled embedding, making it incompatible with our refined concatenation scheme (RC).
Therefore, we do not use $\phi_{a}$ during the training of \ours (of course, it is used in inference). Instead, we focus on training $\phi_s$ while incorporating attribute information in the text domain. Specifically, we generate text triplets using BLIP-generated captions to separately capture subject and attribute information, through ``Efficient In-Context Learning'' (see \cref{subsec:appedix_triplet_details}). Then, we provide subject information to $\phi_s$ while directly using attribute information in text format. For example, given text triplets: $T_r: $ ``a room with a chair and a table'', $T_c: $ ``replace the chair with a sofa'', $T_t: $ ``a room with a sofa and a table''. our refined concatenation scheme (RC) ensures that only textual subject information is input to $\phi_s$. This extraction is feasible because  BLIP-generated captions (from FTI4CIR) already provide subject and attribute separation.

\textbf{Context-I2W} \citep{Context-I2W} introduces a context-dependent projection module that selects relevant visual information. To achieve this, Context-I2W employs a more complex projection mechanism that utilizes context captions such as ``A [REPLACE] passes the ball with his teammate during a training session'', which removes the first subject term. Then, these context captions and full visual sequence embeddings from the visual encoder are fused in the Context-I2W projection module. Since it requires full sequences of visual embeddings rather than a single pooled embedding, it can be incompatible with our refined concatenation scheme (RC).
In such cases, our method can still be applied simply by replacing the text encoder.
Namely, instead of training RTD with this projection module, we simply replace the text encoder with the one updated by \ours using the projection module $\phi$ from Pic2Word \citep{pic2word}. Namely, during inference, we use the projection module from Context-I2W alongside the \ours-updated text encoder (which was trained with the projection module from Pic2Word). The strong performance of this variant suggests that \ours does not need to be trained with each specific projection module, again highlighting its strong generalizability.

\section{Additional experimental results} \label{sec:appendix_additional_experiments}

\subsection{Results on GeneCIS \citep{genecis} and COCO object composition}
\label{subsec:appendix_genecis_and_coco}
We observe that integrating our approach with projection-based CIR methods results in consistent yet marginal performance improvements on GeneCIS, as shown in \cref{tab:genecis}.
The relatively smaller performance difference compared to other datasets can be attributed to the discrepancy between the format of the conditioning text of GeneCIS and the projection-based CIR methods training methodology.
Namely, GeneCIS only uses the fixed four text instructions ``change attribute'', ``focus attribute'', ``change object'' and ``focus object'', which is different from the usual text instruction we expected (\eg, ``change the dog to a cat'').

In the experiment on COCO object composition, we observe a significant performance improvement, similar to the results obtained on other datasets in \cref{tab:coco}.
This finding reaffirms that our approach, when combined with ZS-CIR methods, consistently achieves strong performance, demonstrating its generalizability.

\begin{table}[h!]
    \centering
    \caption{GeneCIS results}
    \resizebox{0.7\columnwidth}{!}{

\input{Table/GeneCIS}
    }
    \label{tab:genecis}
\end{table}
\begin{table}[h!]
    \centering
    \caption{COCO object composition results}
    \resizebox{0.7\columnwidth}{!}{
    \input{Table/coco}
    }
    \label{tab:coco}
\end{table}

\subsection{Results on different backbones  (ViT-B/32, ViT-L/14)} \label{subsec:appendix_different_backbones}
\cref{table:fashioniq} summarizes the evaluation results on the FashionIQ dataset. In the table, we observe that the incorporation of our approach with projection-based CIR methods significantly improves the performance across all three existing projection-based CIR methods (SEARLE, Pic2Word, and LinCIR) and all backbones (ViT-B/32 and ViT-L/14).
For example, regardless of the choice of projection-based CIR methods and backbones, the minimum performance gain for average R@10 and R@50 scores is greater than 2 and 3.5 points, respectively.
\cref{tab:cirrandcirco} shows a similar trend on the CIRR and CIRCO datasets. 
Notably, in some metrics on the CIRR and CIRCO datasets, the performance improvements achieved through our method (ViT-B/32) even exceed those obtained by employing a larger backbone (ViT-L/14), which demonstrates the effect of our method.
Specifically, in the CIRR R@1 score, SEARLE + \ours (26.29) and LinCIR + \ours (24.82) using ViT-B/32 surpasses the original results of SEARLE (24.89)  and LinCIR (23.76) using ViT-L/14.

\begin{table}[t]
    \centering
    \caption{\textbf{FashionIQ validation results.} The results of \ours combined with Pic2Word \citep{pic2word}, SEARLE \citep{searle}, and LinCIR \citep{lincir} across different CLIP backbones (ViT-B/32 and ViT-L/14) are shown.}
    \resizebox{\columnwidth}{!}{
    \setlength{\tabcolsep}{2pt}

\input{Table/FashionIQ}
    }
    \label{table:fashioniq}
\end{table}
\begin{table}[t]
    \centering
    \caption{\textbf{CIRR and CIRCO test results.} Details are the same as \cref{table:fashioniq}.}
    \resizebox{\columnwidth}{!}{
    \setlength{\tabcolsep}{5pt}

\input{Table/CIRRandCIRCO}
    }
    \label{tab:cirrandcirco}
\end{table}

\subsection{Results on larger backbone (ViT-G/14)} \label{subsec:appendix_larger_backbone}
As reported in \cref{tab:appendix_comparison},  we evaluate the performance of \ours using the significantly larger backbone (OpenCLIP ViT-G/14 \citep{openclip}). As described in \cref{subsec:ablation_studies}, we use the projection module $\phi$ from LinCIR \citep{lincir}. 
Since the pre-trained projection module $\phi$ for LinCIR \citep{lincir} (ViT-G/14) is not publicly available, we reproduce it and integrate \ours with it.
We emphasize that similar to our previous results, \ours again achieves remarkable gains across all datasets. Here, we set the learning rate as $10^{-6}$.

\begin{table}[h!]
    \centering
    \caption{FashionIQ results on larger OpenCLIP ViT-G/14 backbone \citep{openclip}.}
    \resizebox{\columnwidth}{!}{

\input{Table/giga_FashionIQ}
    }
    \label{tab:giga_fashionIQ}
\end{table}
\begin{table}[h!]
    \centering
    \caption{CIRR and CIRCO results on larger OpenCLIP ViT-G/14 backbone \citep{openclip}.}
    \resizebox{\columnwidth}{!}{

\input{Table/giga_CIRRandCIRCO}
    }
    \label{tab:giga_CIRRandCIRCO}
\end{table}

\begin{table}[h!]
    \centering
    \caption{\textbf{More efficient variants.} ``Learnable params (\%)'' denotes the percentage of learnable parameters relative to the entire set of parameters in the text encoder. }
    \resizebox{\columnwidth}{!}{

\input{Table/ablation_efficient_variants}
    }
    \label{tab:ablation_efficient_variant}
\end{table}

\subsection{More efficient variants} \label{subsec:more_efficient_variants}
\cref{tab:ablation_efficient_variant} presents the results of the more efficient implementations of our approach in terms of the number of updated parameters. 
Specifically, instead of updating the entire set of parameters of the text encoder, we explore updating only a few layers of the network when applying \ours, 
Our findings indicate that updating only the fully connected layers (denoted as ``Whole FCs'') nearly matches the performance of the full model while using less than half the number of learnable parameters (40.72 vs. 40.53 average scores). 
Additionally, we verify that updating only three fully connected layers, whose parameter size matches the projection module $\phi$ and constitutes 11.5\% of the full model, is also sufficiently effective. 
We test various three-layer updating strategies: ``First 3 FCs'': the first three layers (closest to the input), ``middle 3 FCs'': the middle three layers, ``Last 3 FCs'': the last three layers, and ``Interleave 3 FCs'': an interleaved selection of three layers (first, middle, and last layers).
Among these, we verify that the ``Interleave 3 FCs'' shows the best result, maintaining competitive performance with the full model (40.72 vs. 39.88 average scores).
We believe these findings suggest a promising direction for enhancing the training efficiency of our approach by selectively updating only specific layers of the text encoder.

\subsection{Effectiveness of \ours across dataset scales} \label{subsec:appendix_data_scales}
We conduct experiments with various scales of training text triplets. For the small-scale text triplets, we sub-sampled text triplets from LLM-based text triplets. Thus, the last row in the \cref{tab:data_scale_small} denotes the original result (LLM-based \ours result). We also measure the effectiveness of \ours using large-scale text triplets (up to 5M) by combining publicly available text triplets (IP2P, CoVR) with ours (LLM-based, rule-based). Here, validation splits of all three benchmark datasets are utilized and full results will be included in the final version. 

As shown in \cref{tab:data_scale_small,tab:data_scale_large}, we confirm that small-scale text triplets are sufficient to achieve the effectiveness of \ours. We believe the main reason for this is that, to reduce task discrepancy, only the relationship between the concatenated caption (reference caption + conditioning caption, $T_{r+c}$) and the target caption $T_t$ needs to be learned. We believe this learning task requires much less data compared to learning representations from scratch. Moreover, since the text encoder is already pre-trained, the model does not need significant changes to learn this simple but crucial learning task for CIR.

\begin{table}[h!]
    \centering
    \caption{\textbf{Results across different scales of LLM-based text triplets}. In each row, text triplets are sub-sampled from 2.5M original LLM-based text triplets provided by Compodiff \citep{compodiff}}
    \resizebox{0.8\columnwidth}{!}{
    \input{Table/ablation_data_scale_small}
    }
    \label{tab:data_scale_small}
\end{table}

\begin{table}[h!]
    \centering
    \caption{Results of larger-sized text triplets}
    \resizebox{\columnwidth}{!}{
    \input{Table/ablation_data_scale_large}
    }
    \label{tab:data_scale_large}
\end{table}

\subsection{Ablations on filtering process} \label{subsec:appendix_filtering_experiment}
In rule-based text triplet generation, we highlight that the filtering process using the projection module from LinCIR is \textit{marginally effective}. As demonstrated in the \cref{tab:filtering_ablation}, even without the filtering procedure, the enhancement of \ours from LinCIR remains considerable.  This result demonstrates that the effectiveness of our rule-based text triplets is not solely dependent on the use of the projection module from LinCIR in the filtering process.

\begin{table}[h!]
    \centering
    \caption{Ablations on filtering process}
    \resizebox{\columnwidth}{!}{
    \input{Table/filtering_ablation}
    }
    \label{tab:filtering_ablation}
\end{table}

\begin{table}[ht]
    \centering
    \caption{Noise type variation on CIRR/CIRCO dataset}
    \resizebox{\columnwidth}{!}{
    \input{Table/noise_CIRR_CIRCO}
    }
    \label{tab:noise_CIRR}
\end{table}

\begin{table}[ht]
    \centering
    \caption{Noise type variation on FashionIQ dataset}
    \resizebox{\columnwidth}{!}{
    \input{Table/noise_FashionIQ}
    }
    \label{tab:noise_FashionIQ}
\end{table}

\subsection{Ablations on noise injection} \label{subsec:appendix_noise_injection}
We conduct an ablation study of the different noise types employed for the ``refined concatenation scheme'' shown in \cref{fig:method_diagram}. We compare three different noise types, uniform distribution, Gaussian distribution, and LinCIR-ish noise ($\text{Unif}(0, 1) \times \mathcal N(0, 1) $). 
We also examine the scale of LinCIR-ish noise from 0.1, 0.5, and 1.
We report the test scores for CIRR and CIRCO, as well as the FashionIQ validation scores for Pic2Word, SEARLE, and LinCIR in \cref{tab:noise_CIRR} and \cref{tab:noise_FashionIQ}. 
In the tables, we observe that all noise distributions show decent performance and LinCIR-like noises show slightly better performances than uniform distribution and normal distribution.
We also observe that the different scale choice for the LinCIR-like noise somewhat affects the overall performances. 
In the main experiments, we chose 0.5 for the noise scale, following the observed performance improvements.

\section{Generating text triplets cost} \label{sec:appendix_training_efficiency}
Although generating text triplets is not our main contribution, for comprehensive understanding, we compare the generation time of LLM-based and rule-based approaches. Even when using LLMs, constructing text triplets is significantly more cost-effective than CIR triplets. Specifically, CIR triplets involve: 1) a subsequent, computationally intensive text-to-image generation phase \citep{instructpix2pix, compodiff}, or 2) the availability of image or video datasets along with an additional collection phase for semantically similar images or videos \citep{covr, (CASE)levy2024data}. In contrast, generating text triplets bypasses these resource-heavy steps.
For example, using 8 A100 GPUs, generating 1M text triplets takes 0.1 hours with the rule-based approach and 3.8 hours with the LLM-based approach from Compodiff \citep{compodiff} (OPT-6.7B). As described in \cref{subsec:appedix_triplet_details}, a more efficient text triplet generation method using in-context learning with LLaMA3-8B reduces the generation time to 1.5 hours without the need for fine-tuning. %

Therefore, while generating text triplets with LLMs incurs a higher cost compared to rule-based methods, it is still significantly faster (15 times) than generating CIR triplets (as used in CompoDiff), which utilize the text
generation step as a preliminary phase for subsequent text-to-image generation. Thus, we believe LLM-based generation remains viable, but the rule-based approach is more efficient.

\begin{figure}[ht]
\centering
\includegraphics[width=\textwidth]{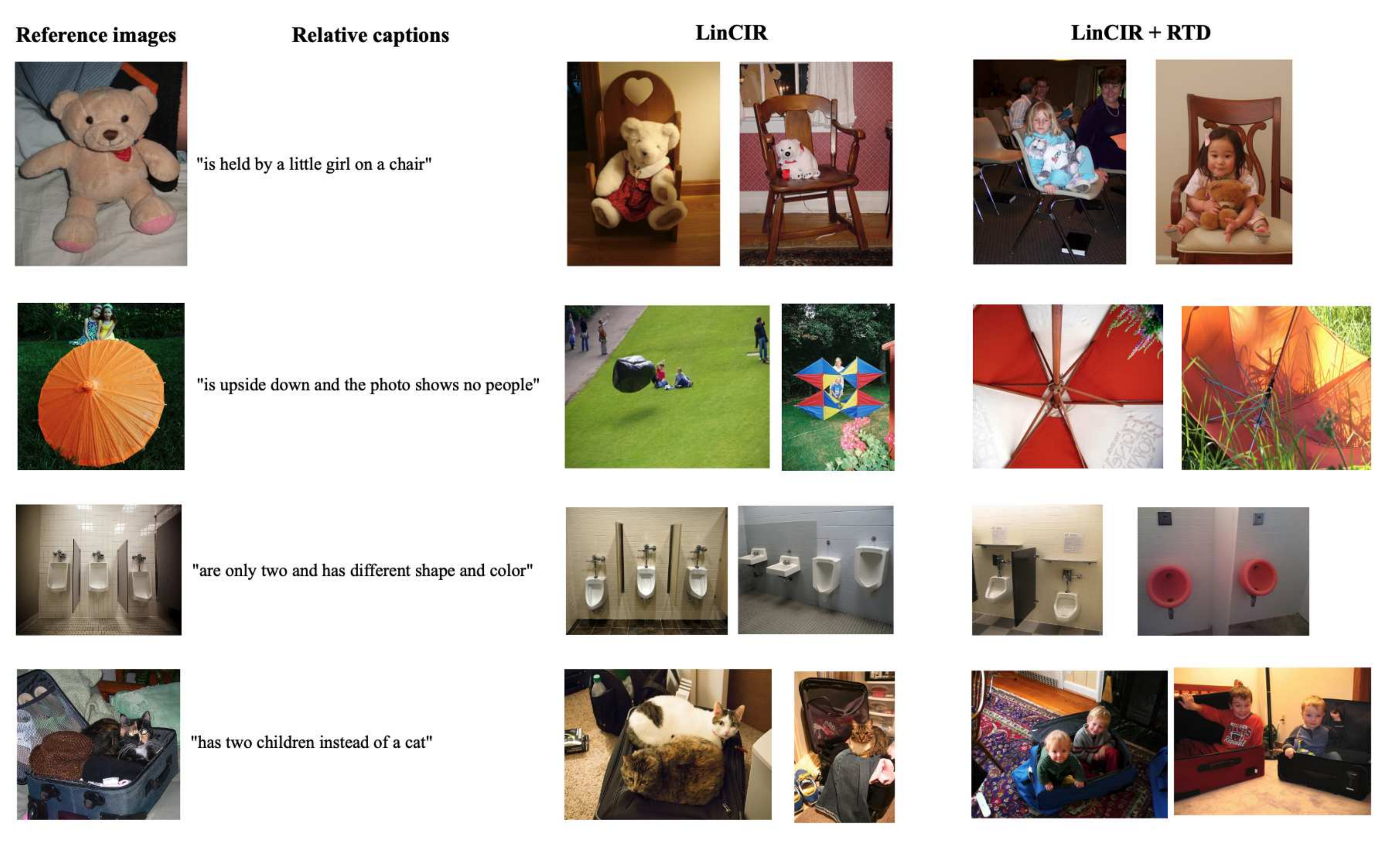} 
\caption{Qualitative Results on CIRCO dataset}
\label{fig:qual}
\vspace{-1em}
\end{figure}

\section{Qualitative example on CIRCO} \label{sec:appendix_qualitative_results}
We qualitatively illustrate the results of incorporating \ours into LinCIR on the CIRCO dataset in \cref{fig:qual}. The visual examples provide an intuitive demonstration of how the integration of \ours enhances the performance of LinCIR, effectively capturing the semantic meaning of the modification descriptions while preserving the relevant visual information from the reference image.

 \section{Discussion and Limitations}
 \label{sec:limitations}
 We have primarily focused on evaluating the integrability of \ours with representative projection-based CIR methods \citep{pic2word,searle,lincir}. However, we have not yet explored or tested the extensibility of \ours to other CIR approaches that achieve strong performance, such as those utilizing human-annotated CIR triplets (supervised) \citep{combiner}, synthetically generated CIR triplets \citep{covr,(CASE)levy2024data,compodiff}, or training-free methods \citep{cirevl}. Given the core motivation behind \ours, its adaptability to those CIR approaches that directly train fusion modules or backbones using CIR triplets may be limited. However, considering the strong performance and practical advantages—such as efficient training and inference—offered by projection-based CIR methods compared to other variants, we believe that integrating \ours with them remains a valuable direction in the CIR domain.

\section{Societal Impacts}\label{sec:societal_impacts}
Although our paper demonstrates promising outcomes in the ZS-CIR task, further examination of the data and the model is essential prior to practical deployment. Since our method focuses mainly on optimization for accuracy, unwanted social implications can occur. For example, real-world images from databases and user-generated text may inadvertently cause harmful cases.

\section{Reproducibility Statement}
We provide all necessary details for reproduction in the manuscript, including implementation details, metrics, datasets, and baselines, as described in \cref{sec:ExpSetup}. Additionally, the training and evaluation dataset details are elaborated in \cref{subsec:appendix_dataset_details,subsec:appedix_triplet_details}. The anonymized code for reproducing our results is provided in the supplementary material.

%% file: Table/rebuttal_language_table_example.tex
\begin{tabular}{cccc}
\toprule
\textbf{\Large Text triplets}          & \textbf{\Large Reference text $T_r$}                     & \textbf{\Large Conditioning text $T_c$}                    & \textbf{\Large Target text $T_t$ }                                \\
\midrule
{\Large Rule-based} & {\Large ``another wall at my home"}                      & {\Large ``bedroom is added in place of wall"}          & {\Large ``another bedroom at my home"}                           \\ 
\midrule
{\Large IP2P \citep{instructpix2pix}}                 & {\Large ``watercolor of your pet!"}                      & {\Large ``make it a huge grizzly bear"}                & {\Large ``watercolor of a huge grizzly bear!"}                   \\
                              
\midrule
{\Large Compodiff \citep{compodiff} }      & {\Large ``Chinese landscape watercolor painting"}        & {\Large ``make the landscape a cityscape"}             & {\Large ``chinese cityscape watercolor painting"}                \\

\midrule
{\Large Efficient in-context learning} & {\Large ``young business woman on a bench"}                      & {\Large ``add a laptop"}          & {\Large ``young business woman on a bench with a laptop"}                           \\
                                 
\midrule

{\Large CoVR \citep{covr}}                 & {\Large ``Two little boys are running"}                    & {\Large ``Have them dance"}                              & {\Large ``Two little boys are dancing"}                            \\
                              
\midrule
{\Large CASE \citep{(CASE)levy2024data}}                & \color{blue}{\Large ``A scone with an orange slice on a plate"}      & \color{blue}{\Large ``This food is not acidic"}                    & \color{blue}{\Large ``a close up of a muffin on a plate on a table"}         \\
                                     
\bottomrule
\end{tabular}

%% file: Table/rebuttal_text_generation_information.tex
\begin{tabular}{lccccc}
\toprule
\multicolumn{1}{c}{\textbf{Dataset}}  & \textbf{Use LLM} & \textbf{Model}  & \textbf{Fine-tuning strategy} &\textbf{\# of text triplets} \\
\midrule
\multicolumn{1}{c}{Rule-based} & \nomark       & \nomark                            & \nomark & 1.3M                     \\
\hline

\multicolumn{1}{c}{IP2P \citep{instructpix2pix}}                                     & \yesmark       & GPT-3                                 & Fine-tuned on 700 human-written text triplets     & 450K     \\

\multicolumn{1}{c}{Compodiff \citep{compodiff}}      & \yesmark       & OPT-6.7B         &   Fine-tuned on 450k IP2P text triplets &2.5M          \\ 
\multicolumn{1}{c}{Efficient in-context learning}      & \yesmark       & LLaMA3-8B         &   In-context learning  & 1M               \\ 

\multicolumn{1}{c}{CoVR \citep{covr}}                                     & \yesmark       & LLaMA-7B                               & Fine-tuned on 700 human-written text triplets  & 700K             \\
\multicolumn{1}{c}{CASE \citep{(CASE)levy2024data}}                                    & \yesmark       & GPT-3                                   & In-context learning & 350K               \\
\bottomrule
\end{tabular}

%% file: Table/template.tex
\begin{tabular}{p{8cm}p{8cm}}
\toprule
"replace \verb|${source}| with \verb|${target}|" & "substitute \verb|${target}| for \verb|${source}|" \\
"apply \verb|${target}|" & "\verb|${source}| is removed and \verb|${target}| takes its place" \\
"convert \verb|${source}| to \verb|${target}|" & "modify \verb|${source}| to become \verb|${target}|" \\
"replace \verb|${source}| with \verb|${target}|" & "customize \verb|${source}| to become \verb|${target}|" \\
"update \verb|${source}| to \verb|${target}|" & "change \verb|${source}| to match \verb|${target}|" \\
"substitute \verb|${target}| for \verb|${source}|" & "\verb|${target}| is introduced after \verb|${source}| is removed" \\
"alter \verb|${source}| to match \verb|${target}|" & "\verb|${target}| is added in place of \verb|${source}|" \\
"upgrade \verb|${source}| to \verb|${target}|" & "\verb|${target}| is introduced as the new option after" \\
"amend \verb|${source}| to fit \verb|${target}|" & "\verb|${source}| is removed and \verb|${target}| is added" \\
"opt for \verb|${target}|" & "\verb|${source}| is removed and \verb|${target}| is introduced" \\
"\verb|${source}| is removed" & "\verb|${target}| is added as a replacement for \verb|${source}|" \\
"add \verb|${target}|" & "\verb|${target}| is the new option available" \\
"if it is \verb|${target}|" & "\verb|${target}| is added after \verb|${source}| is removed" \\
"\verb|${target}| is the updated option" & "\verb|${target}| is introduced after \verb|${source}| is retired" \\
"\verb|${target}| is the updated choice" & "tweak \verb|${source}| to become \verb|${target}|" \\
"\verb|${source}| is replaced with \verb|${target}|" & "has no \verb|${source}|" \\
"change \verb|${source}| to \verb|${target}|" & "alter \verb|${source}| to \verb|${target}|" \\
"swap \verb|${source}| for \verb|${target}|" & "redesign \verb|${source}| as \verb|${target}|" \\
"turn \verb|${source}| into \verb|${target}|" & "adapt \verb|${source}| to fit \verb|${target}|" \\
"choose \verb|${target}| instead of \verb|${source}|" & "\verb|${target}| is the new choice" \\
"\verb|${target}| is the new selection" & "exchange \verb|${source}| with \verb|${target}|" \\
"transform \verb|${source}| into \verb|${target}|" & "show no \verb|${source}|" \\
"no \verb|${source}|" & "remove \verb|${source}|" \\
"delete \verb|${source}|" & "not a \verb|${source}|" \\
"with no \verb|${source}|" & "without \verb|${source}|" \\
\bottomrule
\end{tabular}

%% file: Table/GeneCIS.tex
\begin{tabular}{cclll}
\toprule
& & \multicolumn{3}{c}{Average} \\
& & \multicolumn{1}{c}{R@1} & \multicolumn{1}{c}{R@2} & \multicolumn{1}{c}{R@3} \\
\midrule
\multirow{6}{*}{ViT-B} & Pic2Word & $11.13$ & $21.08$ & $31.05$ \\
& +\ours & $12.03$ \color{violet}{($\mathbf{+0.90}$)} & $21.61$ \color{violet}{($\mathbf{+0.53}$)} & $31.09$ \color{violet}{($\mathbf{+0.04}$)} \\
& SEARLE & $12.19$ & $22.56$ & $32.03$ \\
& +\ours & $12.82$ \color{violet}{($\mathbf{+0.63}$)} & $22.97$ \color{violet}{($\mathbf{+0.41}$)} & $32.44$ \color{violet}{($\mathbf{+0.41}$)} \\
& LinCIR & $12.23$ & $21.29$ & $30.80$ \\
& +\ours & $12.83$ \color{violet}{($\mathbf{+0.60}$)} & $22.83$ \color{violet}{($\mathbf{+1.54}$)} & $32.22$ \color{violet}{($\mathbf{+1.42}$)} \\
\midrule
\multirow{6}{*}{ViT-L} & Pic2Word & $11.18$ & $21.45$ & $30.55$ \\
& +\ours & $11.92$ \color{violet}{($\mathbf{+0.74}$)} & $22.32$ \color{violet}{($\mathbf{+0.87}$)} & $31.33$ \color{violet}{($\mathbf{+0.78}$)} \\
& SEARLE & $12.30$ & $22.08$ & $31.29$ \\
& +\ours & $12.40$ \color{violet}{($\mathbf{+0.10}$)} & $22.82$ \color{violet}{($\mathbf{+0.74}$)} & $32.37$ \color{violet}{($\mathbf{+1.08}$)} \\
& LinCIR & $12.45$ & $22.66$ & $32.06$ \\
& +\ours & $13.18$ \color{violet}{($\mathbf{+0.73}$)} & $23.12$ \color{violet}{($\mathbf{+0.46}$)} & $32.77$ \color{violet}{($\mathbf{+0.71}$)} \\
\bottomrule
\end{tabular}

%% file: Table/coco.tex
\begin{tabular}{cclll}
\toprule
&          & \multicolumn{3}{c}{COCO}                                                                                                            \\
&          & \multicolumn{1}{c}{R@1}                      & \multicolumn{1}{c}{R@5}                      & \multicolumn{1}{c}{R@10}                     \\
\midrule
\multirow{6}{*}{ViT-B} & Pic2Word & $6.88$                                       & $13.6$                                       & $17.52$                                      \\
& +\ours    & $7.62$ \color{violet}{($\mathbf{+0.74}$)}      & $20.23$ \color{violet}{($\mathbf{+6.63}$)}     & $28.69$ \color{violet}{($\mathbf{+11.17}$)}    \\
& SEARLE   & $9.52$                                       & $21.45$                                      & $29.38$                                      \\
& +\ours    & $11.01$ \color{violet}{($\mathbf{+1.49}$)}     & $24.34$ \color{violet}{($\mathbf{+2.89}$)}     & $32.84$ \color{violet}{($\mathbf{+3.46}$)}     \\
& LinCIR   & $7.15$                                       & $18.38$                                      & $27.3$                                       \\
& +\ours    & $9.59$ \color{violet}{($\mathbf{+2.44}$)}      & $21.66$ \color{violet}{($\mathbf{+3.28}$)}     & $30.66$ \color{violet}{($\mathbf{+3.36}$)}     \\
\midrule
\multirow{6}{*}{ViT-L} & Pic2Word & $10.26$                                      & $23.67$                                      & $32.53$                                      \\
& +\ours    & $10.26$ \color{violet}{($\mathbf{+0.00}$)}     & $24.66$ \color{violet}{($\mathbf{+0.99}$)}     & $33.56$ \color{violet}{($\mathbf{+1.03}$)}     \\
& SEARLE   & $12.07$                                      & $26.13$                                      & $35.17$                                      \\
& +\ours    & $14.38$ \color{violet}{($\mathbf{+2.31}$)}     & $29.74$ \color{violet}{($\mathbf{+3.61}$)}     & $38.09$ \color{violet}{($\mathbf{+2.92}$)}     \\
& LinCIR   & $11.37$                                      & $24.53$                                      & $33.85$                                      \\
& +\ours    & $14.6$ \color{violet}{($\mathbf{+3.23}$)}      & $29.84$ \color{violet}{($\mathbf{+5.31}$)}     & $38.87$ \color{violet}{($\mathbf{+5.02}$)}     \\
\bottomrule
\end{tabular}

%% file: Table/FashionIQ.tex
\begin{tabular}{c|cllllllll}
\toprule
& & \multicolumn{2}{c}{\Large{Shirt}} & \multicolumn{2}{c}{\Large{Dress}} & \multicolumn{2}{c}{\Large{Toptee}} & \multicolumn{2}{c}{\Large{Average}} \\
& & \multicolumn{1}{c}{\large{R@10}} & \multicolumn{1}{c}{\large{R@50}} & \multicolumn{1}{c}{\large{R@10}} & \multicolumn{1}{c}{\large{R@50}} & \multicolumn{1}{c}{\large{R@10}} & \multicolumn{1}{c}{\large{R@50}} & \multicolumn{1}{c}{\large{R@10}} & \multicolumn{1}{c}{\large{R@50}} \\
\midrule
\multirow{6}{*}{ViT-B/32} & Pic2Word & 13.40 & 28.46 & 8.48 & 20.77 & 13.31 & 29.68 & 11.73 & 26.30 \\

& +\ours & 23.06 \color{violet}($\mathbf{+9.66}$) & 40.48 \color{violet}($\mathbf{+12.02}$) & 20.33 \color{violet}($\mathbf{+11.85}$) & 41.75 \color{violet}($\mathbf{+20.98}$) & 24.12 \color{violet}($\mathbf{+10.81}$) & 46.35 \color{violet}($\mathbf{+16.67}$) & 22.5 \color{violet}($\mathbf{+10.77}$) & 42.86 \color{violet}($\mathbf{+16.56}$) \\
\cmidrule{2-10}
\\[-2.5mm]
& SEARLE & 24.78 & 41.85 & 17.90 & 36.99 & 25.24 & 46.71 & 22.64 & 41.85 \\
& +\ours & 26.69 \color{violet}($\mathbf{+1.91}$) & 44.31 \color{violet}($\mathbf{+2.46}$) & 20.72 \color{violet}($\mathbf{+2.82}$) & 43.13 \color{violet}($\mathbf{+6.14}$) & 26.67 \color{violet}($\mathbf{+1.43}$) & 48.75 \color{violet}($\mathbf{+2.04}$) & 24.7 \color{violet}($\mathbf{+2.06}$) & 45.4 \color{violet}($\mathbf{+3.55}$) \\
\cmidrule{2-10}
\\[-2.5mm]
& LinCIR & 18.55 & 34.64 & 15.67 & 33.86 & 20.19 & 40.08 & 18.14 & 36.20 \\
& +\ours & 23.65 \color{violet}($\mathbf{+5.10}$) & 42.74 \color{violet}($\mathbf{+8.10}$) & 19.98 \color{violet}($\mathbf{+4.31}$) & 41.75 \color{violet}($\mathbf{+7.89}$) & 24.73 \color{violet}($\mathbf{+4.54}$) & 46.56 \color{violet}($\mathbf{+6.48}$) & 22.79 \color{violet}($\mathbf{+4.65}$) & 43.68 \color{violet}($\mathbf{+7.48}$) \\
\midrule
\multirow{6}{*}{ViT-L/14} & Pic2Word & 26.59 & 42.93 & 21.32 & 43.53 & 28.10 & 48.19 & 25.34 & 44.88 \\
& +\ours & 27.97 \color{violet}($\mathbf{+1.38}$) & 46.96 \color{violet}($\mathbf{+4.03}$) & 23.50 \color{violet}($\mathbf{+2.18}$) & 46.65 \color{violet}($\mathbf{+3.12}$) & 31.31 \color{violet}($\mathbf{+3.21}$) & 53.09 \color{violet}($\mathbf{+4.90}$) & 27.59 \color{violet}($\mathbf{+2.25}$) & 48.90 \color{violet}($\mathbf{+4.02}$) \\
\cmidrule{2-10}
\\[-2.5mm]
& SEARLE & 26.94 & 45.34 & 19.58 & 40.80 & 28.45 & 49.77 & 24.99 & 45.30 \\
& +\ours & 32.63 \color{violet}($\mathbf{+5.69}$) & 50.39 \color{violet}($\mathbf{+5.05}$) & 23.2 \color{violet}($\mathbf{+3.62}$) & 47.25 \color{violet}($\mathbf{+6.45}$) & 32.18 \color{violet}($\mathbf{+3.73}$) & 54.56 \color{violet}($\mathbf{+4.79}$) & 29.34 \color{violet}($\mathbf{+4.35}$) & 50.73 \color{violet}($\mathbf{+5.43}$) \\
\cmidrule{2-10}
\\[-2.5mm]
& LinCIR & 30.42 & 47.99 & 21.86 & 44.77 & 29.98 & 50.38 & 27.42 & 47.71 \\
& +\ours & 32.83 \color{violet}($\mathbf{+2.41}$) & 50.44 \color{violet}($\mathbf{+2.45}$) & 24.49 \color{violet}($\mathbf{+2.63}$) & 48.24 \color{violet}($\mathbf{+3.47}$) & 33.4 \color{violet}($\mathbf{+3.42}$) & 54.56 \color{violet}($\mathbf{+4.18}$) & 30.24 \color{violet}($\mathbf{+2.82}$) & 51.08 \color{violet}($\mathbf{+3.37}$) \\
\bottomrule
\end{tabular}

%% file: Table/CIRRandCIRCO.tex
\begin{tabular}{c|clll llll}
\toprule
\multicolumn{2}{c}{} & \multicolumn{3}{c}{\large{CIRR}} & \multicolumn{4}{c}{\large{CIRCO}} \\
\multicolumn{2}{c}{} & \multicolumn{1}{c}{\large{R@1}} & \multicolumn{1}{c}{\large{R@5}}& \multicolumn{1}{c}{\large{R@10}} & \multicolumn{1}{c}{\large{mAP@5}} & \multicolumn{1}{c}{\large{mAP@10}} & \multicolumn{1}{c}{\large{mAP@25}} & \multicolumn{1}{c}{\large{mAP@50}} \\
\midrule
\multirow{6}{*}{ViT-B/32} & Pic2Word & $13.64$ & $37.45$ & $52.22$ & $2.85$ & $3.24$ & $3.89$ & $4.31$ \\
& +\ours & $23.59$ \color{violet}($\mathbf{+9.95}$) & $51.76$ \color{violet}($\mathbf{+14.31}$) & $65.16$ \color{violet}($\mathbf{+12.94}$) & $6.39$ \color{violet}($\mathbf{+3.54}$) & $6.66$ \color{violet}($\mathbf{+3.42}$) & $7.64$ \color{violet}($\mathbf{+3.75}$) & $8.16$ \color{violet}($\mathbf{+3.85}$) \\
\cmidrule{2-9}

& SEARLE & $23.71$ & $53.3$ & $66.84$ & $8.90$ & $9.42$ & $10.64$ & $11.34$ \\
& +\ours & $26.29$ \color{violet}($\mathbf{+2.58}$) & $56.41$ \color{violet}($\mathbf{+3.11}$) & $69.74$ \color{violet}($\mathbf{+2.90}$) & $11.26$ \color{violet}($\mathbf{+2.36}$) & $12.11$ \color{violet}($\mathbf{+2.69}$) & $13.63$ \color{violet}($\mathbf{+2.99}$) & $14.37$ \color{violet}($\mathbf{+3.03}$) \\
\cmidrule{2-9}

& LinCIR & $18.87$ & $45.66$ & $58.43$ & $6.25$ & $6.74$ & $7.62$ & $8.10$ \\
& +\ours & $24.82$ \color{violet}($\mathbf{+5.95}$) & $53.47$ \color{violet}($\mathbf{+7.81}$) & $66.87$ \color{violet}($\mathbf{+8.44}$) & $8.94$ \color{violet}($\mathbf{+2.69}$) & $9.35$ \color{violet}($\mathbf{+2.61}$) & $10.57$ \color{violet}($\mathbf{+2.95}$) & $11.21$ \color{violet}($\mathbf{+3.11}$) \\
\midrule

\multirow{6}{*}{ViT-L/14} & Pic2Word & $24.22$ & $51.49$ & $64.05$ & $8.27$ & $9.10$ & $10.09$ & $10.75$ \\
& +\ours & $27.86$ \color{violet}($\mathbf{+3.64}$) & $56.24$ \color{violet}($\mathbf{+4.75}$) & $68.48$ \color{violet}($\mathbf{+4.43}$) & $9.13$ \color{violet}($\mathbf{+0.86}$) & $9.63$ \color{violet}($\mathbf{+0.53}$) & $10.68$ \color{violet}($\mathbf{+0.59}$) & $11.27$ \color{violet}($\mathbf{+0.52}$) \\
\cmidrule{2-9}

& SEARLE & $24.89$ & $52.31$ & $65.69$ & $11.62$ & $12.72$ & $14.33$ & $15.13$ \\
& +\ours & $26.63$ \color{violet}($\mathbf{+1.74}$) & $56.17$ \color{violet}($\mathbf{+3.86}$) & $68.96$ \color{violet}($\mathbf{+3.27}$) & $16.53$ \color{violet}($\mathbf{+4.91}$) & $17.89$ \color{violet}($\mathbf{+5.17}$) & $19.77$ \color{violet}($\mathbf{+5.44}$) & $20.68$ \color{violet}($\mathbf{+5.55}$) \\
\cmidrule{2-9}

& LinCIR & $23.76$ & $52.89$ & $66.46$ & $13.00$ & $14.11$ & $15.81$ & $16.68$ \\
& +\ours & $26.63$ \color{violet}($\mathbf{+2.87}$) & $56.17$ \color{violet}($\mathbf{+3.28}$) & $68.96$ \color{violet}($\mathbf{+2.50}$) & $17.11$ \color{violet}($\mathbf{+4.11}$) & $18.11$ \color{violet}($\mathbf{+4.00}$) & $20.06$ \color{violet}($\mathbf{+4.25}$) & $21.01$ \color{violet}($\mathbf{+4.33}$) \\
\bottomrule
\end{tabular}

%% file: Table/giga_FashionIQ.tex
\begin{tabular}{ccllllllll}
\toprule
&\multirow{2}{*}{Method} & \multicolumn{2}{c}{\Large{Shirt}} & \multicolumn{2}{c}{\Large{Dress}} & \multicolumn{2}{c}{\Large{Toptee}} & \multicolumn{2}{c}{\Large{Average}} \\
& & \multicolumn{1}{c}{\large{R@10}} & \multicolumn{1}{c}{\large{R@50}} & \multicolumn{1}{c}{\large{R@10}} & \multicolumn{1}{c}{\large{R@50}} & \multicolumn{1}{c}{\large{R@10}} & \multicolumn{1}{c}{\large{R@50}} & \multicolumn{1}{c}{\large{R@10}} & \multicolumn{1}{c}{\large{R@50}} \\ 
\midrule
 & LinCIR (reported in \cite{lincir}) & 46.76 & 65.11 & 38.08 & 60.88 & 50.48 & 71.09 & 45.11 & 65.69 \\ \hline
& LinCIR (reproduced) & 46.61 & 64.72 & 38.18 & 60.54 & 49.26 & 70.83 & 44.68 & 65.36 \\
& +\ours & 47.20 \color{violet}($\mathbf{+0.59}$) & 66.24 \color{violet}($\mathbf{+1.52}$) & 39.86 \color{violet}($\mathbf{+1.68}$) & 63.01 \color{violet}($\mathbf{+2.47}$) & 51.56 \color{violet}($\mathbf{+2.30}$) & 72.51 \color{violet}($\mathbf{+1.68}$) & 46.21 \color{violet}($\mathbf{+1.54}$) & 67.26 \color{violet}($\mathbf{+1.90}$) \\
\midrule

\end{tabular}

%% file: Table/giga_CIRRandCIRCO.tex
\begin{tabular}{cclll llll}
\toprule
&\multirow{2}{*}{ViT-G} & \multicolumn{3}{c}{\large{CIRR}} & \multicolumn{4}{c}{\large{CIRCO}} \\
& & \multicolumn{1}{c}{\large{R@1}} & \multicolumn{1}{c}{\large{R@5}}& \multicolumn{1}{c}{\large{R@10}} & \multicolumn{1}{c}{\large{mAP@5}} & \multicolumn{1}{c}{\large{mAP@10}} & \multicolumn{1}{c}{\large{mAP@25}} & \multicolumn{1}{c}{\large{mAP@50}} \\
\midrule
& LinCIR (reported in \cite{lincir}) & $35.25$ & $64.72$ & $76.05$ & $19.81$ & $21.01$ & $23.03$ & $24.18$ \\ \hline
& LinCIR (reproduced) & $34.94$ & $64.51$ & $76.12$ & $20.63$ & $21.93$ & $24.12$ & $25.20$ \\ 
& +\ours & $36.31$ \color{violet}($\mathbf{+1.37}$) & $67.47$ \color{violet}($\mathbf{+2.96}$) & $78.31$ \color{violet}($\mathbf{+2.19}$) & $21.08$ \color{violet}($\mathbf{+0.45}$) & $22.29$ \color{violet}($\mathbf{+0.36}$) & $24.46$ \color{violet}($\mathbf{+0.34}$) & $25.44$ \color{violet}($\mathbf{+0.24}$) \\

\midrule

\end{tabular}

%% file: Table/ablation_efficient_variants.tex
\begin{tabular}{lcccccccc}
\toprule
& & \multicolumn{2}{c}{CIRR} & \multicolumn{2}{c}{CIRCO} & \multicolumn{2}{c}{FashionIQ} & \multirow{2}{*}{Avg} \\
\multicolumn{1}{c}{Training variants} & \shortstack{Learnable \\ params (\%)}  & R@5 & R@10 & mAP@10 & mAP@25 & R@10 & R@50 &  \\
\midrule
\multicolumn{1}{c}{Baseline(LinCIR) }& 0\% & 54.29 & 67.76 & 12.67 & 14.45 & 27.42 & 47.71 & \multicolumn{1}{c}{37.38} \\  \hline
\\[-3mm]
+\ours (Full model) & 100\% & 57.90 & 71.13 & 16.10 & 17.84 & 30.24 & 51.08 & 40.72 \\ \hline  \\[-3mm]
+\ours (Whole FCs) & 45.8\% & 57.76 & 71.35 & 15.03 & 16.90 & 30.31 & 51.81 & 40.53 \\  \hline   \\[-3mm]
+\ours (Front 3 FCs) & 11.5\% & 55.65 & 69.83 & 13.95 & 15.81 & 28.69 & 49.92 & 38.98 \\ 
+\ours (Middle 3 FCs) & 11.5\% & 56.69 & 70.03 & 14.66 & 16.58 & 28.55 & 49.84 & 39.39 \\
+\ours (Last 3 FCs) & 11.5\% & 56.84 & 69.74 & 14.81 & 16.70 & 29.16 & 50.43 & 39.61 \\
+\ours (Interleave 3 FCs) & 11.5\% & 57.21 & 70.65 & 15.20 & 17.13 & 28.91 & 50.17 & 39.88 \\
\bottomrule
\end{tabular}

%% file: Table/ablation_data_scale_small.tex
\begin{tabular}{c|c|c|c|c}
\hline
\# of triplets & CIRR R@5 & CIRCO mAP@10 & FashionIQ R@10 & Avg  \\
\hline
1K   & 56.64 & 15.66 & 29.89 & 34.06 \\
50K   & 57.40 & 15.95 & 30.77 & 34.71 \\
100K  & 57.16 & 16.03 & 30.57 & 34.59 \\
2.5M  & 57.90 & 16.10 & 30.24 & 34.75 \\
\hline
\end{tabular}

%% file: Table/ablation_data_scale_large.tex
\begin{tabular}{c|c|c|c|c|c|c|c|c}
\hline
IP2P & CoVR & Compodiff & Template-based & CIRR R@5 & CIRCO mAP@10 & FashionIQ R@10 & Avg & \# of triplets \\
\hline
\yesmark &         &          &             & 58.65 & 15.94 & 29.62 & 34.74 & 450k \\
          & \yesmark &          &             & 59.82 & 15.35 & 29.58 & 34.92 & 700k \\
          &         & \yesmark &             & 57.90 & 16.10 & 30.24 & 34.75 & 2.5M \\
          &         &          & \yesmark   & 56.71 & 15.01 & 30.37 & 34.03 & 1.3M \\
\yesmark & \yesmark &          &             & 59.32 & 16.10 & 30.81 & 35.41 & 1.25M \\
\yesmark & \yesmark & \yesmark &             & 59.08 & 16.15 & 30.97 & 35.40 & 3.75M \\
\yesmark & \yesmark & \yesmark & \yesmark  & 58.65 & 16.54 & 31.22 & 35.47 & 5.05M \\
\hline
\end{tabular}

%% file: Table/filtering_ablation.tex
\begin{tabular}{c|c|c|c|c|c}
\hline
\textbf{Type} & \textbf{LinCIR-based filtering} & \textbf{CIRR R@5} & \textbf{CIRCO mAP@10} & \textbf{FashionIQ R@10} & \textbf{Avg} \\ \hline
LinCIR & - & 54.29 & 12.67 & 27.42 & 31.46 \\ 
+RTD (rule-based) & \nomark & 55.49 & 14.75 & 30.24 & 33.49 \\
+RTD (rule-based) & \yesmark & 56.71 & 15.01 & 30.37 & 34.03 \\ \hline
\end{tabular}

%% file: Table/noise_CIRR_CIRCO.tex
\begin{tabular}{ccccrrrrrrr}
\toprule
                                                & \multicolumn{1}{c}{}                        &                      &       & \multicolumn{3}{c}{CIRR}                                                     & \multicolumn{4}{c}{CIRCO}                                                                                                     \\
                                                & \multicolumn{1}{c}{}                        & Noise type           & Scale & \multicolumn{1}{c}{R@1} & \multicolumn{1}{c}{R@5} & \multicolumn{1}{c}{R@10} & \multicolumn{1}{c}{mAP@5}     & \multicolumn{1}{c}{mAP@10}    & \multicolumn{1}{c}{mAP@25}    & \multicolumn{1}{c}{mAP@50}    \\
\midrule
                                               & \multicolumn{1}{c}{Pic2Word} & - & - & 13.64 & 37.45 & 52.22 & 2.85 & 3.24 & 3.89 & 4.31 \\
\cline{2-11}
\\[-3mm]
& & \text{Unif}(-1,1) & 1 & 23.23 & 50.55 & 64.28 & 4.29 & 4.57 & 5.19 & 5.57 \\
& & $\mathcal{N}(0,1)$ & 1 & 21.18 & 47.78 & 61.47 & 4.09 & 4.26 & 4.83 & 5.17 \\
& & $\mathcal{N}(0,1)$ $\times$ \text{Unif}(0,1) & 0.1 & 23.52 & 51.13 & 64.53 & 5.13 & 5.46 & 6.17 & 6.62 \\
& & $\mathcal{N}(0,1)$ $\times$ \text{Unif}(0,1) & 0.5 & 23.01 & 51.18 & 64.84 & 4.29 & 4.57 & 5.19 & 5.57 \\
& \multirow{-5}{*}{+\ours} & $\mathcal{N}(0,1)$ $\times$ \text{Unif}(0,1) & 1 & 23.59 & 51.76 & 65.16 & 6.39 & 6.66 & 7.64 & 8.16 \\
\cline{2-11}
\\[-3mm]
                                                & \multicolumn{1}{c}{SEARLE}                  & -                    & -     & 23.71                   & 53.3                    & 66.84                    & 8.9                           & 9.42                          & 10.64                         & 11.34                         \\
                                                   \cline{2-11}
                                                \\[-3mm]
                                                & \multicolumn{1}{c}{}                        & \text{Unif}(-1,1)              & 1     & 26.07                   & 55.98                   & 69.18                    & 10.87                         & 11.55                         & 12.97                         & 13.65                         \\
                                                & \multicolumn{1}{c}{}                        & $\mathcal{N}(0,1)$             & 1     & 26.41                   & 56.68                   & 69.47                    & 10.91                         & 11.53                         & 12.88                         & 13.6                          \\
                                                & \multicolumn{1}{c}{}                        & $\mathcal{N}(0,1)$ $\times$ \text{Unif}(0,1) & 0.1   & 26.02                   & 55.47                   & 68.15                    & 10.43                         & 11.07                         & 12.37                         & 13.07                         \\
                                                & \multicolumn{1}{c}{}                        & $\mathcal{N}(0,1)$ $\times$ \text{Unif}(0,1) & 0.5   & 26.29                   & 56.41                   & 69.74                    & 11.26                         & 12.11                         & 13.63                         & 14.37                         \\
                                                & \multicolumn{1}{c}{\multirow{-5}{*}{+\ours}} & $\mathcal{N}(0,1)$ $\times$ \text{Unif}(0,1) & 1     & 26.43                   & 56.58                   & 69.76                    & 11.42                         & 12.04                         & 13.38                         & 14.1                          \\
                                                   \cline{2-11}
                                                \\[-3mm]
                                                & \multicolumn{1}{c}{LinCIR}                  & -                    & -     & 18.87                   & 45.66                   & 58.43                    & 6.25                          & 6.74                          & 7.62                          & 8.1                           \\
                                                   \cline{2-11}
                                                \\[-3mm]
                                                & \multicolumn{1}{c}{}                        & \text{Unif}(-1,1)              & 1     & 24.39                   & 52.77                   & 66.39                    & 6.81                          & 7.27                          & 8.28                          & 8.84                          \\
                                                & \multicolumn{1}{c}{}                        & $\mathcal{N}(0,1)$             & 1     & 24.63                   & 53.52                   & 66.63                    & 7.6                           & 7.97                          & 8.92                          & 9.49                          \\
                                                & \multicolumn{1}{c}{}                        & $\mathcal{N}(0,1)$ $\times$ \text{Unif}(0,1) & 0.1   & 24.58                   & 53.3                    & 66.65                    & 9.6                           & 10.11                         & 11.47                         & 12.15                         \\
                                                & \multicolumn{1}{c}{}                        & $\mathcal{N}(0,1)$ $\times$ \text{Unif}(0,1) & 0.5   & 24.82                   & 53.47                   & 66.87                    & 8.94                          & 9.35                          & 10.57                         & 11.21                         \\
\multirow{-18}{*}{ViT-B/32}                     & \multicolumn{1}{c}{\multirow{-5}{*}{+\ours}} & $\mathcal{N}(0,1)$ $\times$ \text{Unif}(0,1) & 1     & 25.4                    & 54.58                   & 67.69                    & 8.17                          & 8.53                          & 9.72                          & 10.35                         \\
\midrule
\multicolumn{1}{l}{}                            & \multicolumn{1}{c}{Pic2Word}                & -                    & -     & 24.22                   & 51.49                   & 64.05                    & 8.27                          & 9.1                           & 10.09                         & 10.75                         \\
   \cline{2-11}
                                                \\[-3mm]
\multicolumn{1}{l}{}                            &                                             & \text{Unif}(-1,1)              & 1     & 28.24                   & 55.95                   & 68.77                    & 8.14                          & 8.81                          & 9.83                          & 10.37                         \\
\multicolumn{1}{l}{}                            &                                             & $\mathcal{N}(0,1)$             & 1     & 27.06                   & 53.95                   & 66.43                    & 7.08                          & 7.66                          & 8.57                          & 9.07                          \\
\multicolumn{1}{l}{}                            &                                             & $\mathcal{N}(0,1)$ $\times$ \text{Unif}(0,1) & 0.1   & 28.24                   & 57.35                   & 68.65                    & 10.04                         & 10.63                         & 11.71                         & 12.31                         \\
\multicolumn{1}{l}{}                            &                                             & $\mathcal{N}(0,1)$ $\times$ \text{Unif}(0,1) & 0.5   & 27.86                   & 56.24                   & 68.48                    & 9.13                          & 9.63                          & 10.68                         & 11.27                         \\
\multicolumn{1}{l}{}                            & \multirow{-5}{*}{+\ours}                     & $\mathcal{N}(0,1)$ $\times$ \text{Unif}(0,1) & 1     & 27.71                   & 55.68                   & 68.02                    & 8.14                          & 8.78                          & 9.84                          & 10.35                         \\
   \cline{2-11}
                                                \\[-3mm]
\multicolumn{1}{l}{}                            & \multicolumn{1}{c}{SEARLE}                  & -                    & -     & 24.89                   & 52.31                   & 65.69                    & 11.62                         & 12.72                         & 14.33                         & 15.13                         \\
   \cline{2-11}
                                                \\[-3mm]
\multicolumn{1}{l}{}                            &                                             & \text{Unif}(-1,1)              & 1     & 26.96                   & 56.99                   & 69.52                    & 15.82 & 16.78 & 18.54 & 19.39 \\
\multicolumn{1}{l}{}                            &                                             & $\mathcal{N}(0,1)$             & 1     & 27.66                   & 57.54                   & 69.57                    & 15.24                         & 15.93                         & 17.65                         & 18.44                         \\
\multicolumn{1}{l}{}                            &                                             & $\mathcal{N}(0,1)$ $\times$ \text{Unif}(0,1) & 0.1   & 26.31                   & 55.88                   & 69.4                     & 16.05                         & 17.26                         & 19.12                         & 20.01                         \\
\multicolumn{1}{l}{}                            &                                             & $\mathcal{N}(0,1)$ $\times$ \text{Unif}(0,1) & 0.5   & 27.04                   & 56.82                   & 69.95                    & 16.53                         & 17.89                         & 19.77                         & 20.68                         \\
\multicolumn{1}{l}{}                            & \multirow{-5}{*}{+\ours}                     & $\mathcal{N}(0,1)$ $\times$ \text{Unif}(0,1) & 1     & 27.93                   & 57.76                   & 70.19                    & 17.35                         & 18.66                         & 20.52                         & 23.44                         \\
   \cline{2-11}
                                                \\[-3mm]
\multicolumn{1}{l}{}                            & \multicolumn{1}{c}{LinCIR}                  & -                    & -     & 23.76                   & 52.89                   & 66.46                    & 13                            & 14.11                         & 15.81                         & 16.68                         \\
   \cline{2-11}
                                                \\[-3mm]
\multicolumn{1}{l}{}                            &                                             & \text{Unif}(-1,1)              & 1     & 26.58                   & 56.31                   & 68.94                    & 17.23                         & 18.2                          & 20.11                         & 21.03                         \\
\multicolumn{1}{l}{}                            &                                             & $\mathcal{N}(0,1)$             & 1     & 26.75                   & 55.64                   & 68.48                    & 16.45                         & 17.57                         & 19.37                         & 20.3                          \\
\multicolumn{1}{l}{}                            &                                             & $\mathcal{N}(0,1)$ $\times$ \text{Unif}(0,1) & 0.1   & 26.7                    & 56.22                   & 69.08                    & 17.24                         & 18.27                         & 20.24                         & 21.19                         \\
\multicolumn{1}{l}{}                            &                                             & $\mathcal{N}(0,1)$ $\times$ \text{Unif}(0,1) & 0.5   & 26.63                   & 56.17                   & 68.96                    & 17.11                         & 18.11                         & 20.06                         & 21.01                         \\
\multicolumn{1}{l}{\multirow{-18}{*}{ViT-L/14}} & \multirow{-5}{*}{+\ours}                     & $\mathcal{N}(0,1)$ $\times$ \text{Unif}(0,1) & 1     & 26.99                   & 56.1                    & 69.01                    & 17.33                         & 18.3                          & 20.21                         & 21.13                         \\
\bottomrule
\end{tabular}

%% file: Table/noise_FashionIQ.tex
\begin{tabular}{ccccrrrrrrrr}
\toprule
\multicolumn{1}{c}{}      & \multicolumn{1}{c}{}                       &                      &       & \multicolumn{2}{c}{Shirt}                           & \multicolumn{2}{c}{Dress}                           & \multicolumn{2}{c}{Toptee}                          & \multicolumn{2}{c}{Average}                         \\
\multicolumn{1}{c}{}      & \multicolumn{1}{c}{}                       & Noise type           & Scale & \multicolumn{1}{c}{R@10} & \multicolumn{1}{c}{R@50} & \multicolumn{1}{c}{R@10} & \multicolumn{1}{c}{R@50} & \multicolumn{1}{c}{R@10} & \multicolumn{1}{c}{R@50} & \multicolumn{1}{c}{R@10} & \multicolumn{1}{c}{R@50} \\
\midrule
& \multicolumn{1}{c}{Pic2Word} & - & - & 13.4 & 28.46 & 8.48 & 20.77 & 13.31 & 29.68 & 11.73 & 26.3 \\
\cline{2-12}
\\[-3mm]
& \multirow{5}{*}{+\ours} & \text{Unif}(-1,1) & 1 & 21.84 & 37.63 & 18.49 & 39.61 & 23.0 & 43.91 & 21.11 & 40.38 \\
& & $\mathcal{N}(0,1)$ & 1 & 20.36 & 37.54 & 16.16 & 38.18 & 21.67 & 42.48 & 19.4 & 39.4 \\
\multicolumn{1}{c}{} & & $\mathcal{N}(0,1)$ $\times$ \text{Unif}(0,1) & 0.1 & 22.23 & 39.35 & 19.98 & 41.7 & 23.81 & 45.23 & 22.01 & 42.09 \\

& & $\mathcal{N}(0,1)$ $\times$ \text{Unif}(0,1) & 0.5 & 24.53 & 43.82 & 20.33 & 41.55 & 26.01 & 48.75 & 23.62 & 44.7 \\
& & $\mathcal{N}(0,1)$ $\times$ \text{Unif}(0,1) & 1 & 23.06 & 40.48 & 20.33 & 41.75 & 24.12 & 46.35 & 22.5 & 42.86 \\
\cline{2-12}
\\[-3mm]

\multicolumn{1}{c}{}      & \multicolumn{1}{c}{SEARLE}                 & -                    & -     & 24.78                    & 41.85                    & 17.90                    & 36.99                    & 25.24                    & 46.71                    & 22.64                    & 41.85                    \\
\cline{2-12}
                                                \\[-3mm]
\multicolumn{1}{c}{}      & \multicolumn{1}{c}{\multirow{5}{*}{+\ours}} & \text{Unif}(-1,1)              & 1     & 23.75                    & 42.25                    & 20.18                    & 40.36                    & 25.14                    & 46.46                    & 23.02                    & 43.02                    \\
                          & \multicolumn{1}{c}{}                       & $\mathcal{N}(0,1)$             & 1     & 24.14                    & 42.25                    & 20.23                    & 40.16                    & 24.17                    & 46.35                    & 22.85                    & 42.92                    \\
                          & \multicolumn{1}{c}{}                       & $\mathcal{N}(0,1)$ $\times$  \text{Unif}(0,1) & 0.1   & 25.12                    & 44.85                    & 20.92                    & 41.40                    & 26.57                    & 47.63                    & 24.20                    & 44.62                    \\
                          & \multicolumn{1}{c}{}                       & $\mathcal{N}(0,1)$ $\times$  \text{Unif}(0,1) & 0.5   & 26.69                    & 44.31                    & 20.72                    & 43.13                    & 26.67                    & 48.75                    & 24.70                    & 45.40                    \\
                          & \multicolumn{1}{c}{}                       & $\mathcal{N}(0,1)$ $\times$  \text{Unif}(0,1) & 1     & 25.07                    & 44.01                    & 20.43                    & 41.00                    & 26.11                    & 47.12                    & 23.87                    & 44.04                    \\
                          \cline{2-12}
                                                \\[-3mm]
\multicolumn{1}{c}{}      & \multicolumn{1}{c}{LinCIR}                 & -                    & -     & 18.55                    & 34.64                    & 15.67                    & 33.86                    & 20.19                    & 40.08                    & 18.14                    & 36.20                    \\
\cline{2-12}
                                                \\[-3mm]
\multicolumn{1}{c}{}      & \multicolumn{1}{c}{\multirow{5}{*}{+\ours}} & \text{Unif}(-1,1)              & 1     & 21.79                    & 39.35                    & 18.89                    & 40.21                    & 23.66                    & 45.33                    & 21.45                    & 41.63                    \\

                          & \multicolumn{1}{c}{}                       & $\mathcal{N}(0,1)$             & 1     & 22.37                    & 38.67                    & 19.53                    & 40.11                    & 23.71                    & 44.37                    & 21.87                    & 41.05                    \\
                          & \multicolumn{1}{c}{}                       & $\mathcal{N}(0,1)$ $\times$  \text{Unif}(0,1) & 0.1   & 23.95                    & 44.11                    & 19.83                    & 41.99                    & 26.62                    & 47.58                    & 23.47                    & 44.56                    \\
                          & \multicolumn{1}{c}{}                       & $\mathcal{N}(0,1)$ $\times$  \text{Unif}(0,1) & 0.5   & 23.65                    & 42.74                    & 19.98                    & 41.75                    & 24.73                    & 46.56                    & 22.79                    & 43.68                    \\
                          \multirow{-18}{*}{ViT-B/32} & \multicolumn{1}{c}{}                       & $\mathcal{N}(0,1)$ $\times$  \text{Unif}(0,1) & 1     & 22.82                    & 41.12                    & 19.78                    & 41.70                    & 25.09                    & 47.07                    & 22.56                    & 43.29                    \\
\midrule
                  & \multicolumn{1}{c}{Pic2Word}               & -                    & -     & 26.59                    & 42.93                    & 21.32                    & 43.53                    & 28.10                    & 48.19                    & 25.34                    & 44.88                    \\
                  \cline{2-12}
                                                \\[-3mm]
                          & \multirow{5}{*}{+\ours}                     & \text{Unif}(-1,1)              & 1     & 27.87                    & 45.93                    & 23.90                    & 46.80                    & 31.21                    & 52.22                    & 27.66                    & 48.32                    \\
                          &                                            & $\mathcal{N}(0,1)$             & 1     & 26.94                    & 44.95                    & 23.45                    & 45.56                    & 30.34                    & 51.45                    & 26.91                    & 47.32                    \\
                          &                                            & $\mathcal{N}(0,1)$ $\times$  \text{Unif}(0,1) & 0.1   & 28.26                    & 47.64                    & 24.05                    & 47.20                    & 31.21                    & 53.70                    & 27.84                    & 49.51                    \\
                          &                                            & $\mathcal{N}(0,1)$ $\times$  \text{Unif}(0,1) & 0.5   & 27.97                    & 46.96                    & 23.50                    & 46.65                    & 31.31                    & 53.09                    & 27.59                    & 48.90                    \\
                          &                                            & $\mathcal{N}(0,1)$ $\times$  \text{Unif}(0,1) & 1     & 28.41                    & 46.91                    & 24.10                    & 46.21                    & 31.11                    & 52.27                    & 27.87                    & 48.46                    \\
                          \cline{2-12}
                                                \\[-3mm]
                          & \multicolumn{1}{c}{SEARLE}                 & -                    & -     & 26.94                    & 45.34                    & 19.58                    & 40.80                    & 28.45                    & 49.77                    & 24.99                    & 45.30                    \\
                          \cline{2-12}
                                                \\[-3mm]
                          & \multirow{5}{*}{+\ours}                     & \text{Unif}(-1,1)              & 1     & 30.13                    & 46.57                    & 22.16                    & 46.90                    & 28.76                    & 50.74                    & 27.02                    & 48.07                    \\
                          &                                            & $\mathcal{N}(0,1)$             & 1     & 26.99                    & 43.23                    & 21.17                    & 44.82                    & 27.54                    & 49.06                    & 25.23                    & 45.70                    \\
                          &                                            & $\mathcal{N}(0,1)$ $\times$  \text{Unif}(0,1) & 0.1   & 32.63                    & 50.39                    & 23.20                    & 47.25                    & 32.18                    & 54.56                    & 29.34                    & 50.73                    \\
                          &                                            & $\mathcal{N}(0,1)$ $\times$  \text{Unif}(0,1) & 0.5   & 31.80                    & 49.31                    & 23.20                    & 47.30                    & 31.41                    & 54.00                    & 28.80                    & 50.20                    \\
                          &                                            & $\mathcal{N}(0,1)$ $\times$  \text{Unif}(0,1) & 1     & 30.03                    & 47.06                    & 22.41                    & 47.05                    & 30.39                    & 52.42                    & 27.61                    & 48.84                    \\
                          \cline{2-12}
                                                \\[-3mm]
                          & \multicolumn{1}{c}{LinCIR}                 & -                    & -     & 30.42                    & 47.99                    & 21.86                    & 44.77                    & 29.98                    & 50.38                    & 27.42                    & 47.71                    \\
                          \cline{2-12}
                                                \\[-3mm]
                          & \multirow{5}{*}{+\ours}                     & \text{Unif}(-1,1)              & 1     & 31.94                    & 50.10                    & 24.44                    & 48.19                    & 33.04                    & 54.26                    & 29.81                    & 50.85                    \\
                          &                                            & $\mathcal{N}(0,1)$             & 1     & 31.70                    & 49.41                    & 23.90                    & 48.19                    & 33.23                    & 53.54                    & 29.27                    & 50.38                    \\
                          &                                            & $\mathcal{N}(0,1)$ $\times$  \text{Unif}(0,1) & 0.1   & 32.92                    & 50.64                    & 24.49                    & 48.74                    & 33.50                    & 55.02                    & 30.31                    & 51.47                    \\
                          &                                            & $\mathcal{N}(0,1)$ $\times$  \text{Unif}(0,1) & 0.5   & 32.83                    & 50.44                    & 24.49                    & 48.24                    & 33.40                    & 54.56                    & 30.24                    & 51.08                    \\
                 \multirow{-18}{*}{ViT-L/14}          &                                            & $\mathcal{N}(0,1)$ $\times$  \text{Unif}(0,1) & 1     & 32.43                    & 50.54                    & 24.64                    & 48.79                    & 33.25                    & 54.77                    & 30.11                    & 51.36                    \\
\bottomrule
\end{tabular}